\documentclass[final, twoside]{IEEEtran} % regular paper
\usepackage{float}
\usepackage{amssymb, amsmath, amsthm, bm}
\usepackage{enumitem}

\makeatletter
\newcommand{\tpmod}[1]{{\@displayfalse\pmod{#1}}}
\makeatother
\newcommand{\var}[1]{\mathit{#1}}
\newcommand{\iu}{{i\mkern1mu}}
\usepackage{graphicx, color}
\graphicspath{{figures-pdf/}}
\usepackage{cite}
\usepackage{url}
\usepackage{diagbox}
\usepackage[aboveskip=1pt]{subcaption}
\usepackage{pbox}
\usepackage[mathlines]{lineno}
\usepackage{enumitem}
\usepackage{algorithm}
\usepackage{algpseudocode}
\usepackage{multirow, bigstrut}
\usepackage{tabularx}
\usepackage{arydshln}
\usepackage{empheq}
\usepackage{threeparttable}
\newlength\OneImW
\newlength\BigOneImW
\newlength\twofigwidth
\newlength\ThreeImW
\newlength\sfigwidth
\newlength\vfigskip
\newlength\figsep
\newtheorem{Property}{Property}

\newcommand{\x}[0]{{\bf x}}

\usepackage[bookmarks=false]{hyperref}
\hypersetup{
 linktocpage=true, pdfborderstyle={/S/S/W 1}, hyperindex=true, bookmarks=true, bookmarksopen=true, bookmarksnumbered=true, 
}
\begin{document}
	
\title{Adjusting Dynamics of Hopfield Neural Network via Time-variant Stimulus}

\author{Xuenan Peng, Chengqing Li, Yicheng Zeng, Chun-Lai Li
%~\IEEEmembership{Senior Member, ~IEEE, }
% <-this % stops a space
\thanks{This work was supported by the National Natural Science Foundation of China (no.~92267102).}

\thanks{X. Peng is with the School of Mathematics and Computational Science, Xiangtan University, Xiangtan 411105, Hunan, China.}

\thanks{C. Li is with the Key Laboratory of Intelligent Computing \& Information Processing of the Ministry of Education, Xiangtan University, Xiangtan 411105, Hunan, China (DrChengqingLi@gmail.com).}

\thanks{Y. Zeng is with the School of Physics and Optoelectronic Engineering, Xiangtan University, Xiangtan, 411105, Hunan, China.}

\thanks{C.-L. Li is with the School of Computer Science, Xiangtan University, Xiangtan, 411105, Hunan, China.}
}

% The paper headers
\markboth{IEEE Transactions}{Peng \MakeLowercase{et al.}}
	
% put a publisher's ID mark on the page
\IEEEpubid{\begin{minipage}{\textwidth}\ \\[12pt] \centering
 1549-8328 \copyright 2019 IEEE. Personal use is permitted, but republication/redistribution requires IEEE permission.\\
 See http://www.ieee.org/publications standards/publications/rights/index.html for more information.
\end{minipage}}

\maketitle

\begin{abstract}
As a paradigmatic model for nonlinear dynamics studies, the Hopfield Neural Network (HNN) demonstrates a high susceptibility to external disturbances owing to its intricate structure. This paper delves into the challenge of modulating HNN dynamics through time-variant stimuli.  The effects of adjustments using two distinct types of time-variant stimuli, namely the Weight Matrix Stimulus (WMS) and the State Variable Stimulus (SVS), along with a Constant Stimulus (CS) are reported. The findings reveal that deploying four WMSs enables the HNN to generate either a four-scroll or a coexisting two-scroll attractor. When combined with one SVS, four WMSs can lead to the formation of an eight-scroll or four-scroll attractor, while the integration of four WMSs and multiple SVSs can induce grid-multi-scroll attractors. Moreover, the introduction of a CS and an SVS can significantly disrupt the dynamic behavior of the HNN. Consequently, suitable adjustment methods are crucial for enhancing the network's dynamics, whereas inappropriate applications can lead to the loss of its chaotic characteristics.
To empirically validate these enhancement effects, the study employs an FPGA hardware platform.  Subsequently, an image encryption scheme is designed to demonstrate the practical application benefits of the dynamically adjusted HNN in secure multimedia communication.  This exploration into the dynamic modulation of HNN via time-variant stimuli offers insightful contributions to the advancement of secure communication technologies.
\end{abstract} 
\begin{IEEEkeywords} 
Adjustable dynamics, chaotic attractor, chaotic encryption, 
Hopfield neural network, secure communication, time-variant stimulus.
\end{IEEEkeywords}

\section{Introduction}

\IEEEPARstart{N}{eural} networks endeavor to mimic the behavioral attributes of biological neural systems through sophisticated mathematical models \cite{Sung:synaptic:NatureC22}. Various models have been formulated, including the Fitzhugh-Nagumo \cite{Carletti:FNnetwork:PRE20}, Hodgkin-Huxley \cite{Shama:HHnetwork:TNSRE20, Haghiri:HHnetwork:TCSI21}, Morris-Lecar \cite{Hayati:MLnetwork:TCSI15}, Hindmarsh-Rose \cite{Lu:HRmodel:NC23}, and Hopfield Neural Network (HNN) \cite{Masaki:Hopfield:TNNL22}. These models are instrumental in exploring the impact of neuronal interactions, simulating the intricate structure and functionality of the human brain.
Neural network models are broadly classified into two categories based on their architecture: feed-forward and recurrent neural networks. Feed-forward networks are characterized by the absence of cyclic paths within their structure, whereas recurrent networks incorporate connections between nodes that create a directed, temporal graph \cite{Adly:Hopfield:TM22}. This architectural distinction renders recurrent networks particularly effective in modeling temporal dynamics and emulating biological memory processes.

As a prototypical example of a recurrent neural network, the HNN exhibits exceptional dynamics due to its nonlinear activation function and complex network structure \cite{Lai:Hopfield:TNNLS22}. This network finds diverse applications in pathology and biology \cite{Lin:Hopfield:TCSII23, Yu:Hopfield:TNSE23, Bao:Hopfield:TIE23}. The intriguing phenomenon of frustrated chaos in the HNN was examined through bifurcation diagrams, particularly concerning the variation in connection weights \cite{Hugues:chaos:NN2002}. Introducing nonlinear synaptic weights in a four-dimensional HNN model led to the emergence of multiple attractors and the concurrent existence of re-merging Feigenbaum trees \cite{Njitacke:dynamics:AEU2018}.
In efforts to further enhance the dynamic capabilities of the HNN, various nonlinear synapse simulators, like memristors, have been incorporated to replace traditional constant synaptic weights. For example, the application of a flux-controlled memristor as a synaptic element in a four-neuron HNN yielded multi-scroll chaotic attractors \cite{Lai:memristive:TCSI23}. Similarly, integrating three memristor-based synapses into an HNN unveiled complex multi-structure attractors and initial-offset behaviors \cite{Lin:memristive:TCDICS23}. A noteworthy discovery was a hidden attractor in a memristive synapse-based HNN, characterized by non-intersecting basins of attraction near any unstable equilibrium point \cite{Lin:meristor:TCSII20}.
Subsequent studies uncovered a pair of hidden attractors in a three-neuron HNN \cite{Danca:hidden:CSF2017}, highlighting various dynamic behaviors such as bursting oscillation and coexisting attractors with offset boosting in different HNN configurations \cite{Valle:dynamics:TNNLS2018, Chen:dynamics:Neucom2020, Zhang:dynamics:ND2020}. Despite these advances, tailoring an HNN to achieve specific outcomes in alternative network topologies remains a formidable challenge.

\IEEEpubidadjcol % must call \IEEEpubidadjcol in the second column for its text to clear the %IEEEpubid mark. 

The inherent instability of the HNN makes it particularly sensitive to external disturbances, a characteristic that spurred researchers to test it in varied environments to elicit a spectrum of dynamic behaviors \cite{Wu:electromagnetic:AMC2019}. For example, Lin's exploration of the HNN under electromagnetic conditions revealed complex phenomena such as hyper-chaos and transient chaos \cite{Lin:electromagnetic:ND2020}. In another study, an HNN demonstrated multi-stability dynamics when subjected to an electromagnetic field, verified through breadboard experimentation \cite{Li:HNNmultistability:TCSII22}. These findings prompted further research into placing the HNN in even more intricate settings involving multiple radiations, resulting in a range of dynamic responses \cite{Wan:Hopfield:chaos22}.
Additional factors like noise, time delays, and external stimuli were found to significantly influence the behavior of HNN \cite{Hu:time-delay:AMC2019, Lu:noise:ND2019, Evaldo:stimulus:Physa2021}. Among these, time-variant stimuli emerge as a particularly effective tool for modulating neural network dynamics due to their easy implementation and control. Lin's study highlighted the emergence of chaotic dynamics in an HNN subjected to both electromagnetic radiation and time-variant stimuli \cite{Lin:stimulus:CNSNS2020}. Notably, in these configurations, the HNN tends to maintain or extend its attractors primarily in one direction.

In this study, the effects of two time-variant stimuli, namely Weight Matrix Stimulus (WMS) and State Variable Stimulus (SVS), along with a Constant Stimulus (CS), on the dynamics of an HNN are investigated. This exploration results in four distinct models, each characterized by different combinations of stimuli: four WMSs, four WMSs with one SVS, four WMSs with multiple SVSs, and a combination of one CS with one SVS. The dynamic outcomes of these models are diverse, including a four-scroll or coexisting two-scroll attractor, a four-scroll or eight-scroll attractor, and a grid multi-scroll attractor. Particularly notable is the final model where the Adjusted HNN (AHNN) exhibits a range of behaviors from chaotic to quasi-periodic, periodic, and ultimately fixed points as the intensity of the CS is varied. This indicates that the appropriate application of stimuli can significantly enhance the dynamics of the HNN, whereas inappropriate methods may lead to its degeneration. Further, the AHNN is encoded by Verilog HDL for implementation on an FPGA platform. To aid in real-world applications, a piecewise Taylor series is employed to approximate the hyperbolic tangent function. Lastly, an image encryption scheme is proposed that utilizes the high randomness of the AHNN, with its time-variant parameters serving as the secret key.

The remainder of the paper is organized as follows: In Section~\ref{sec:models}, four adjusted versions of HNN are established with three stimuli.
Section~\ref{sec:dynamics} delves into the analysis of the dynamics of the four types of AHNN. The implementation of AHNN on the FPGA platform is demonstrated in Section~\ref{sec:FPGA}. Subsequently, an image encryption algorithm utilizing AHNN is detailed in Section~\ref{sec:algorithm}.
The last section concludes the paper.

\section{The model of Adjusted Hopfield neural network}
\label{sec:models}

\begin{figure}[!htb]
\begin{minipage}{\twofigwidth}
\centering
 \includegraphics[width=0.95\twofigwidth]{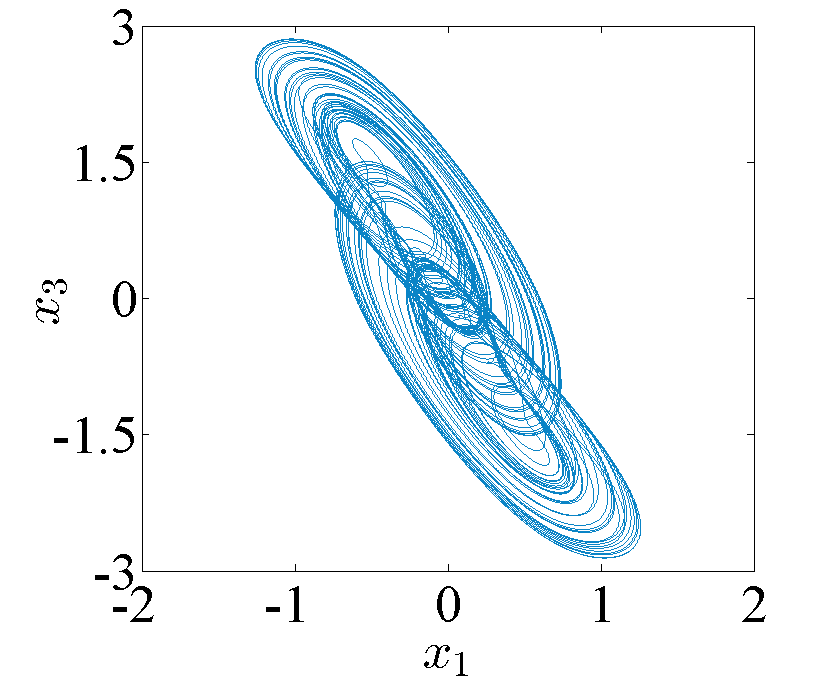}\\
(a)
\end{minipage}
\begin{minipage}{\twofigwidth}
\centering
 \includegraphics[width=0.95\twofigwidth]{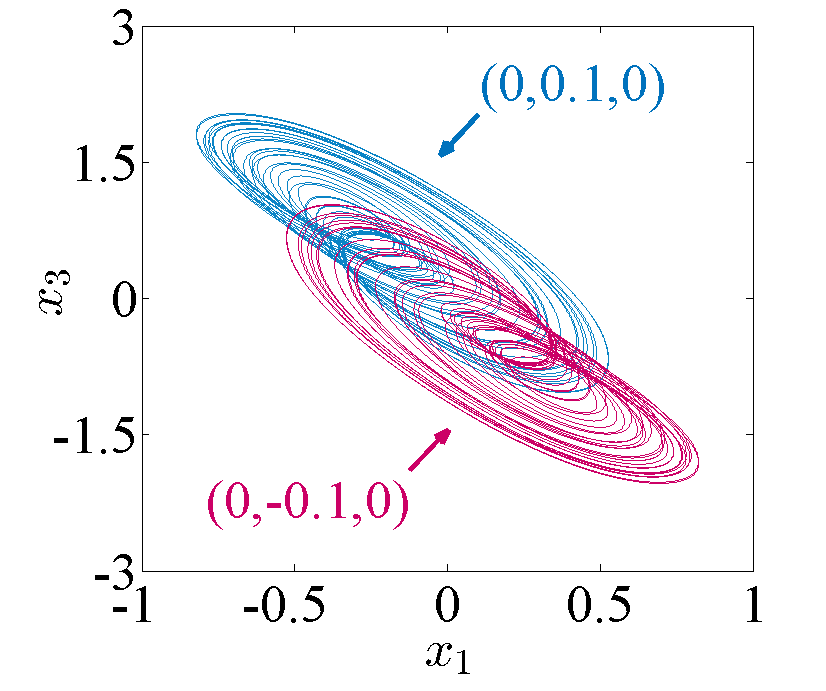}\\
(b)
\end{minipage}
\caption{Phase diagram of network~\eqref{eq:OHNN}:
(a) two-scroll attractors; (b) coexisting one-scroll attractors.}
\label{fig:Oattractor}
\end{figure}

In an HNN consisting of $n$ interconnected neurons \cite{Tang:HNN-stimuli:ND24}, the dynamics of the $i$-th neuron in relation to the others is
\begin{equation}
C_i\frac{dx_i}{dt}=-\frac{x_i}{R_i}+\sum_{j=1}^n\omega_{ij}\tanh{(x_j)}+I_i,
\end{equation}
where $x_{i}$ denotes the voltage across the membrane capacitor $C_{i}$, $R_{i}$ signifies the resistance of the cell membrane, $\tanh(x_{j})$ represents the activation function, $\omega_{ij}$ is the synaptic weight between the $i$-th and $j$-th neurons, and $I_{i}$ is the external current excitation. A specific case of an HNN with three coupled neurons is 
\begin{equation}
\centering
\dot{\x}= -\x +
\begin{bmatrix}
2.2 & -1.2 & 0.5 \\
2 & 1.5 & k \\
-5 & 0 & -1
\end{bmatrix}
\cdot
\begin{bmatrix}
\tanh(x_1)\\
\tanh(x_2)\\
\tanh(x_3)
\end{bmatrix},
\label{eq:OHNN}
\end{equation}
where $k \in \{1, 1.15\}$ and $\x=[x_1, x_2, x_3]^\mathrm{T}$. Figure~\ref{fig:Oattractor}(a) displays a two-scroll chaotic attractor in the network~\eqref{eq:OHNN} for $k=1.15$ with an initial condition of $(x_{1}(0), x_{2}(0), x_{3}(0))=(0, 0.1, 0)$. Conversely, for $k=1$ and initial conditions $(x_{1}(0), x_{2}(0), x_{3}(0))\in \{(0, 0.1, 0), (0, -0.1, 0) \}$, the network~\eqref{eq:OHNN} generates a pair of coexisting attractors, as depicted in Fig.~\ref{fig:Oattractor}(b).

\begin{figure}[!htb]
\centering
\includegraphics[width=0.8\BigOneImW]{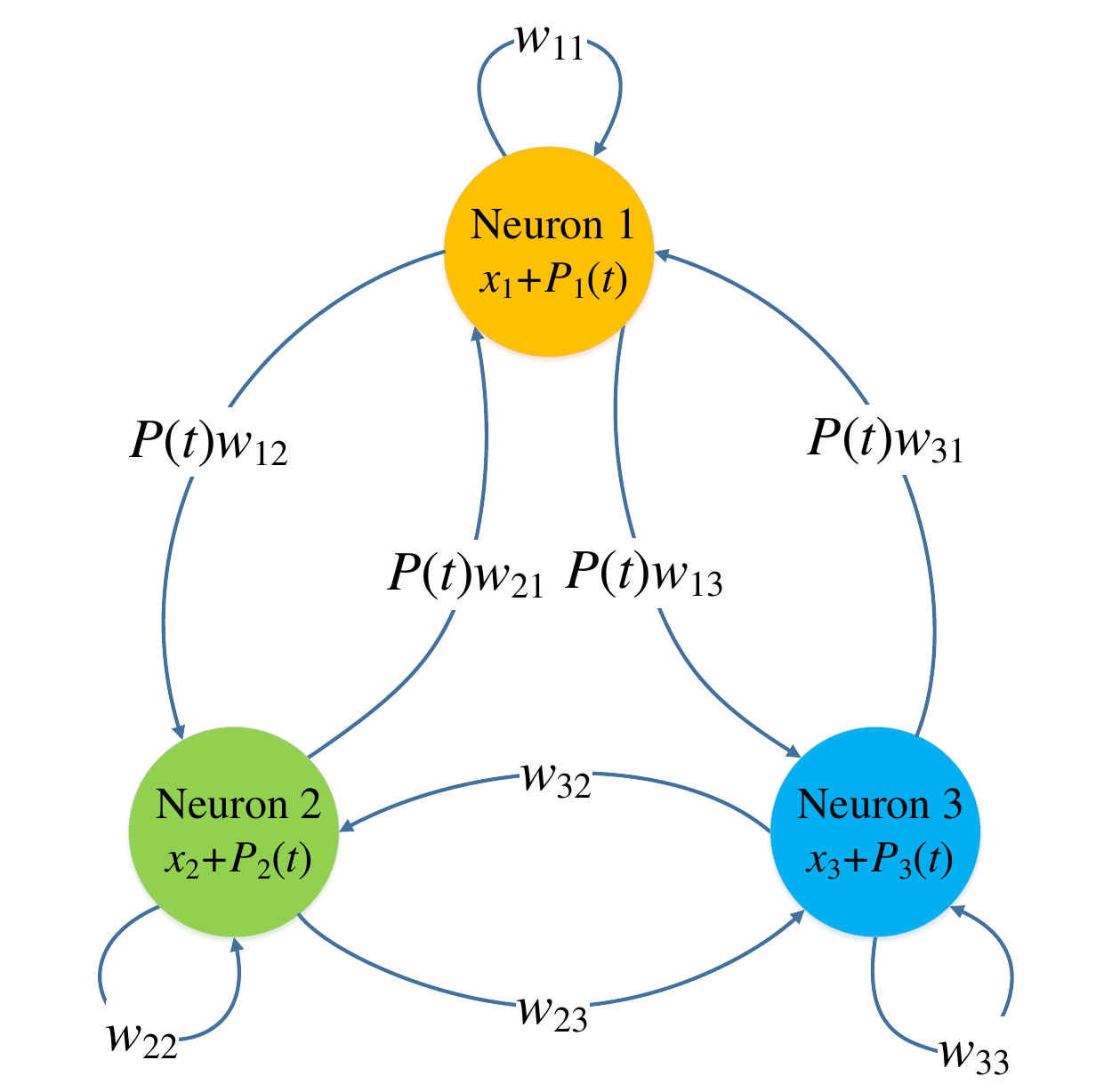}
\caption{The structure of AHNN with time-variant stimuli.}
\label{fig:structure}
\end{figure}

Two distinct types of time-variant stimuli are
\begin{equation}
P(t) = A \cdot \text{sign}(\sin(\omega t))
\end{equation}
and
\begin{equation}
P_i(t) = \frac{A_i}{2}(\text{sign}(\sin(\omega_i t)) + 1).
\end{equation}
These stimuli, referred to as WMS and SVS respectively, are governed by their angular velocities ($\omega$ and $\omega_i$) and amplitudes ($A$ and $A_i$) \cite{Wu:pulse:Chaos2019, Hong:pulse:TCAD2019}. Each stimulus operates using a binary logic, where $P(t)$ oscillates between $\{-A, A\}$ and $P_i(t)$ alternates between $\{0, A_i\}$.
Subsequently, these stimuli are applied to the network~\eqref{eq:OHNN}, as depicted in Figure~\ref{fig:structure}. The mathematical framework of the AHNN is
\begin{equation}
    \dot{\mathbf{x}} =-
    \begin{bmatrix}
      x_1 + P_1(t) \\
      x_2 + P_2(t) \\
      x_3 + P_3(t)
    \end{bmatrix}
    +
    \mathbf{W} \cdot
    \begin{bmatrix}
      \tanh (x_1 + P_1(t)) \\
      \tanh (x_2 + P_2(t)) \\
      \tanh (x_3 + P_3(t))
    \end{bmatrix},
    \label{eq:HNNstimuli}
\end{equation}
where
\begin{equation}
    \mathbf{W} =
    \begin{bmatrix}
      2.2    & -1.2P(t) & 0.5P(t) \\
      2P(t)  & 1.5      & k \\
      -5P(t) & 0        & -1
    \end{bmatrix}.
\end{equation}

\begin{figure}[!htb]
 \begin{minipage}{\twofigwidth}
 \centering
 \includegraphics[width=\twofigwidth]{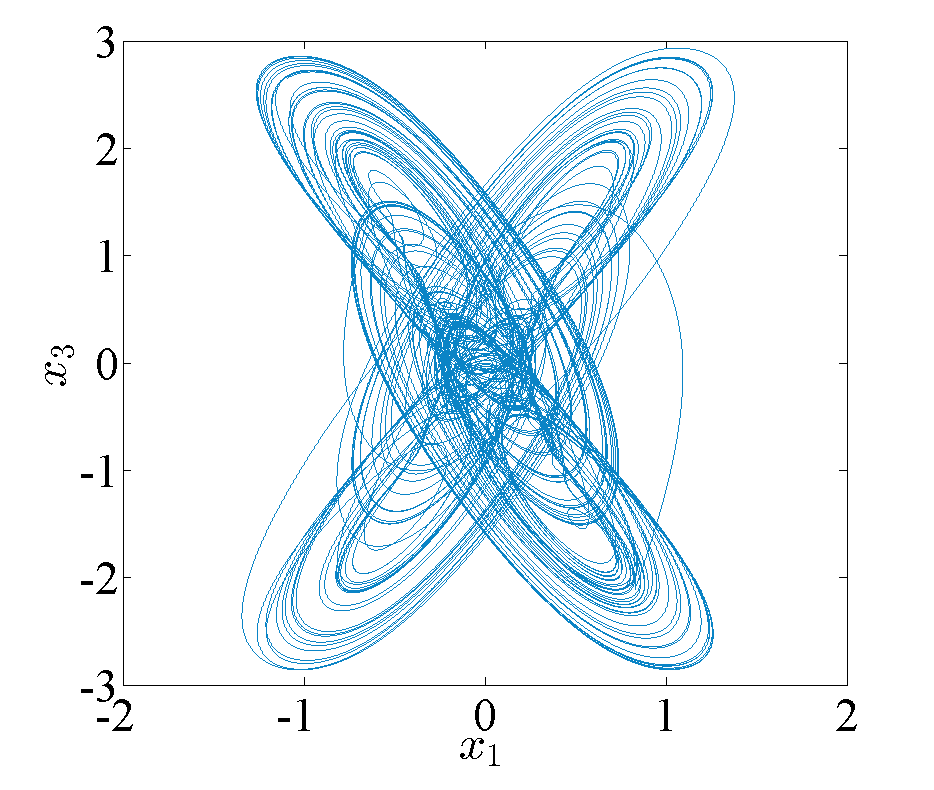}\\
 (a)
\end{minipage}
\begin{minipage}{\twofigwidth}
 \centering
 \includegraphics[width=\twofigwidth]{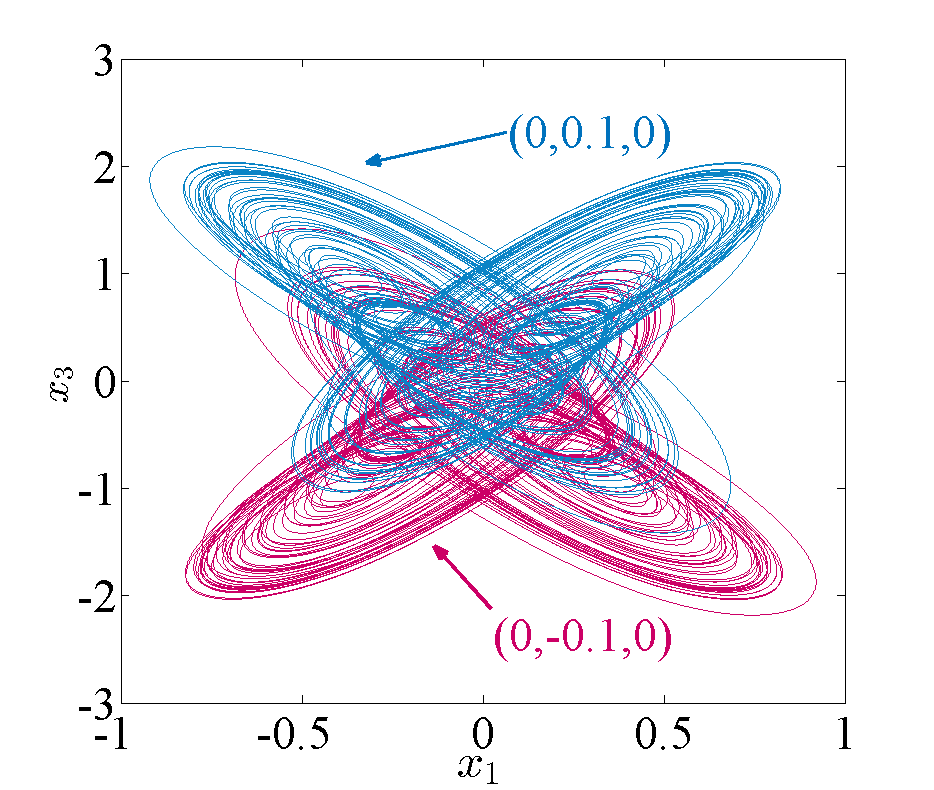}\\
 (b)
\end{minipage}   
\caption{Phase diagram of AHNN with four WMSs:
(a) four-scroll chaotic attractor; (b) two-scroll coexisting attractors.}
\label{fig:WMSattractor}
\end{figure}

\begin{figure}[!htb]
 \centering
 \begin{minipage}{\ThreeImW}
  \centering
  \includegraphics[width=0.95\ThreeImW]{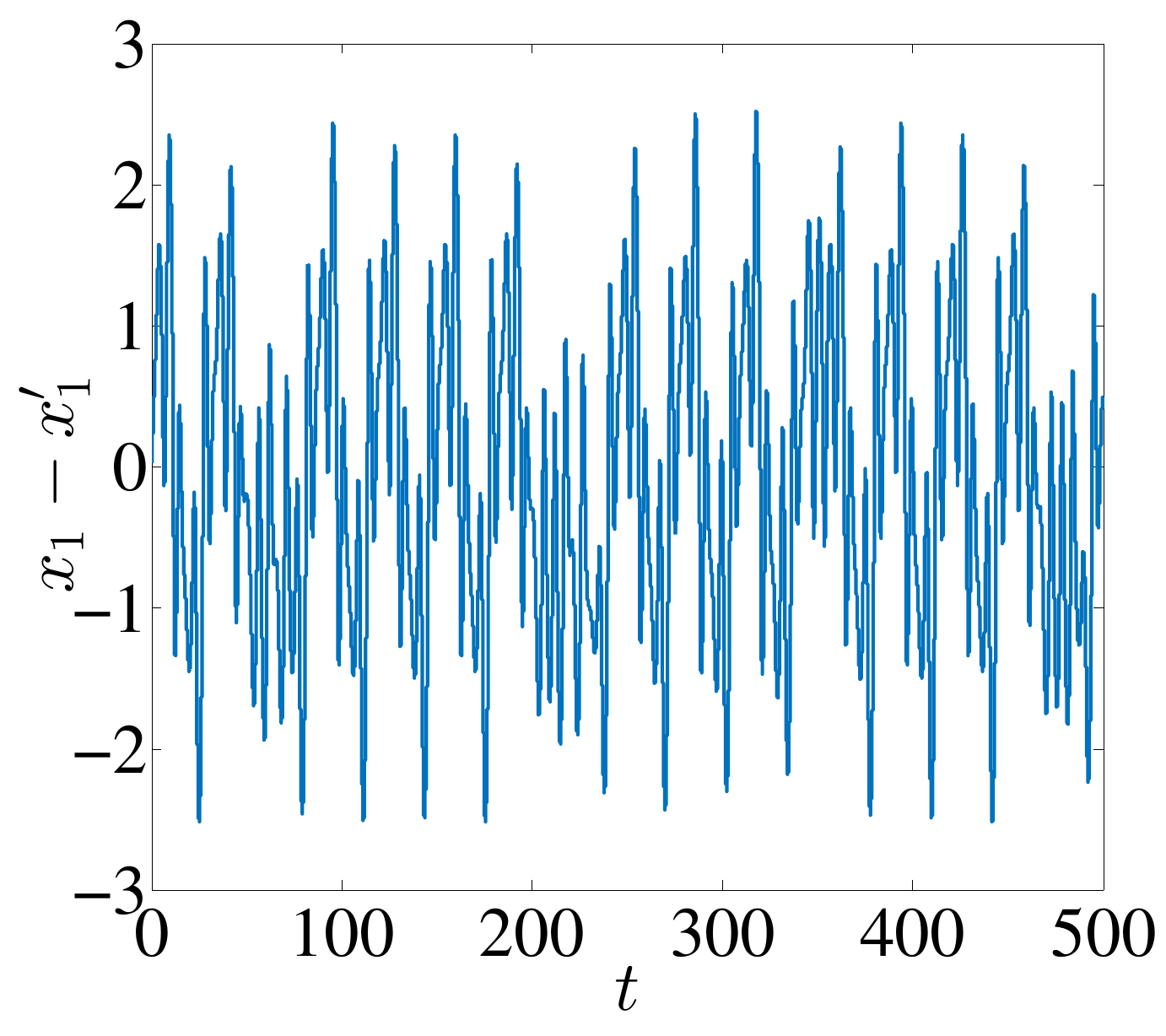} \\
  (a)
 \end{minipage}
 \begin{minipage}{\ThreeImW}
  \centering
  \includegraphics[width=\ThreeImW]{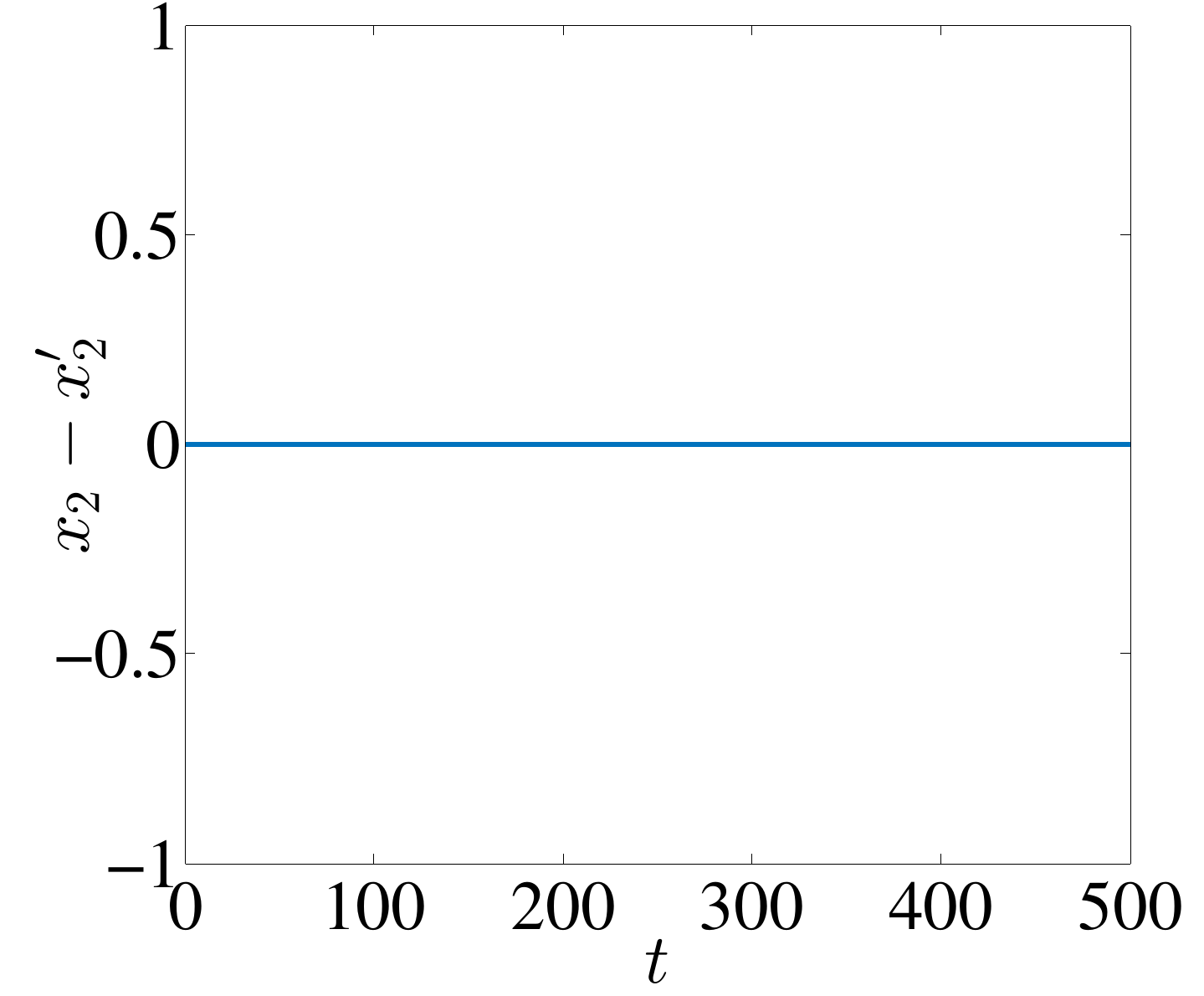}\\
  (b)
 \end{minipage}
 \begin{minipage}{\ThreeImW}
  \centering
  \includegraphics[width=\ThreeImW]{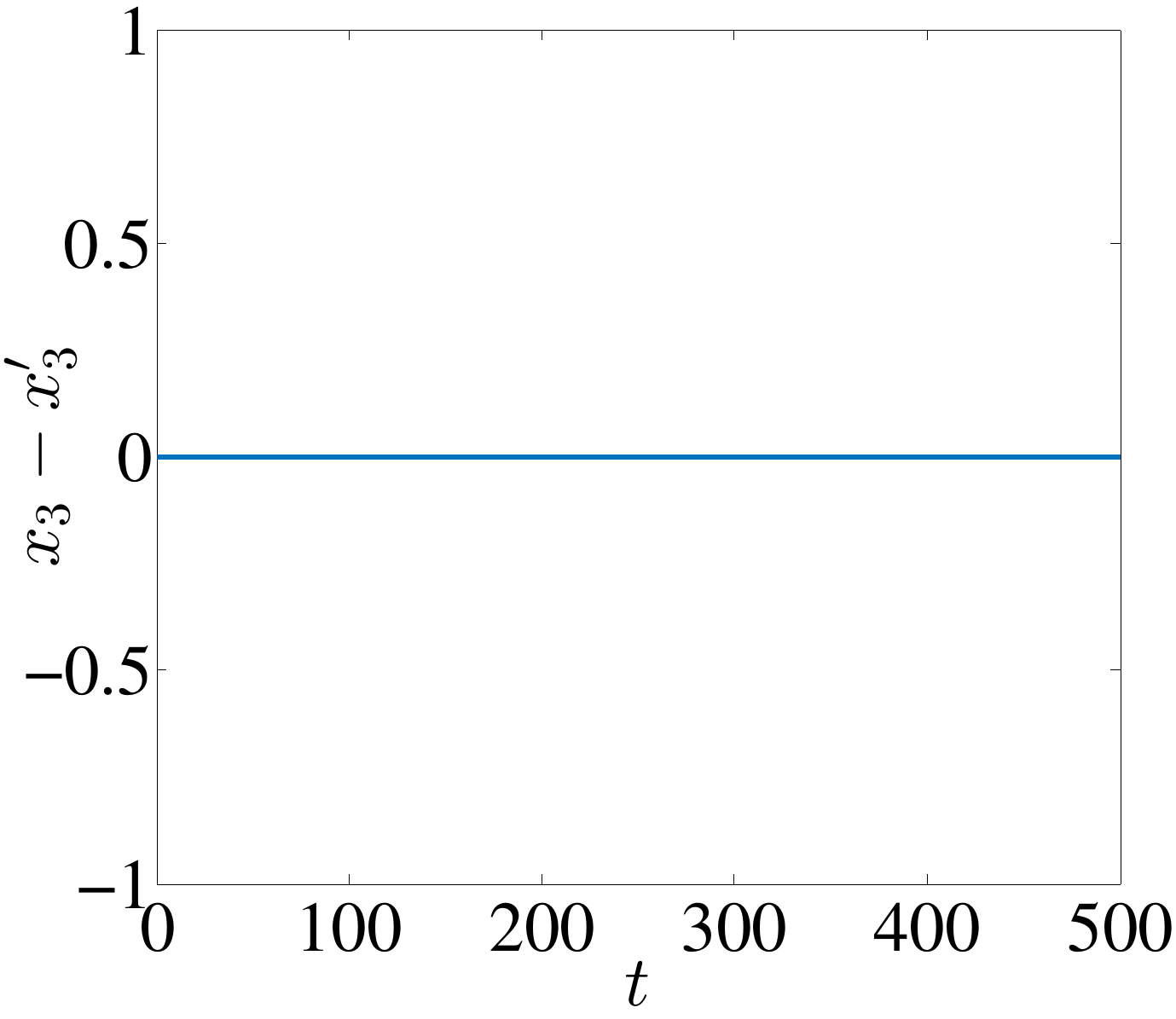}\\
  (c)
 \end{minipage}
 \caption{The time series of AHNN with $P(t) = \pm 1$: (a) $x_1-x'_1$; (b) $x_2-x'_2$; (c) $x_3-x'_3$.}
\label{fig:WMStime}
\end{figure}

\begin{figure}[!htb]
 \centering
 \begin{minipage}{\twofigwidth}
  \centering
  \includegraphics[width=\twofigwidth]{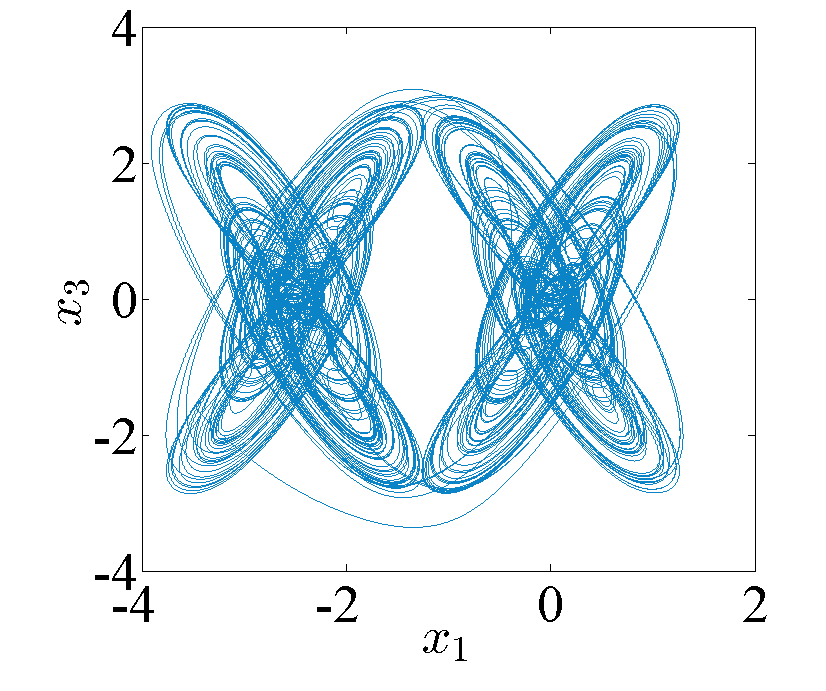}\\
  (a)
 \end{minipage}
 \begin{minipage}{\twofigwidth}
  \centering
  \includegraphics[width=\twofigwidth]{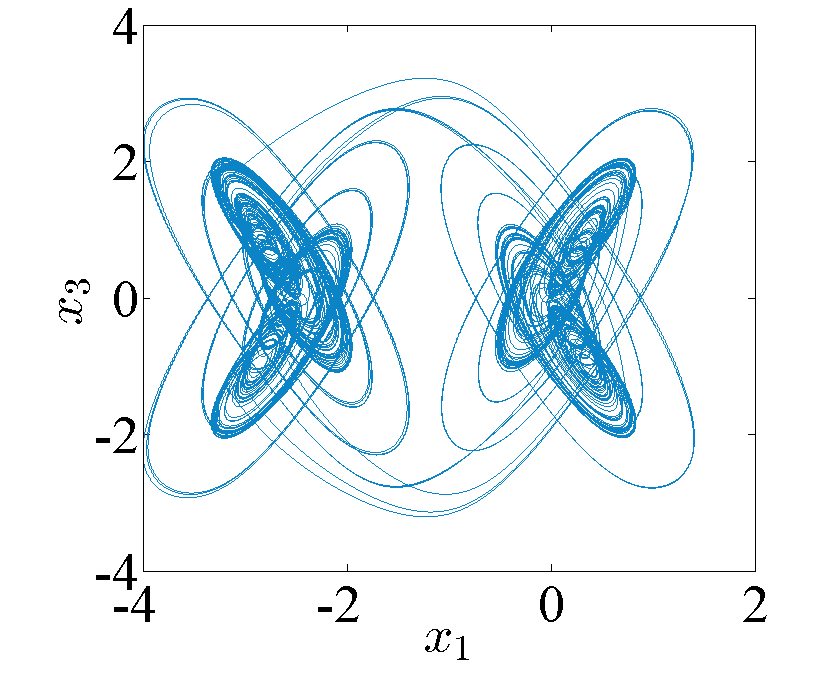}\\
  (b)
 \end{minipage}
  \begin{minipage}{\twofigwidth}
  \centering
  \includegraphics[width=\twofigwidth]{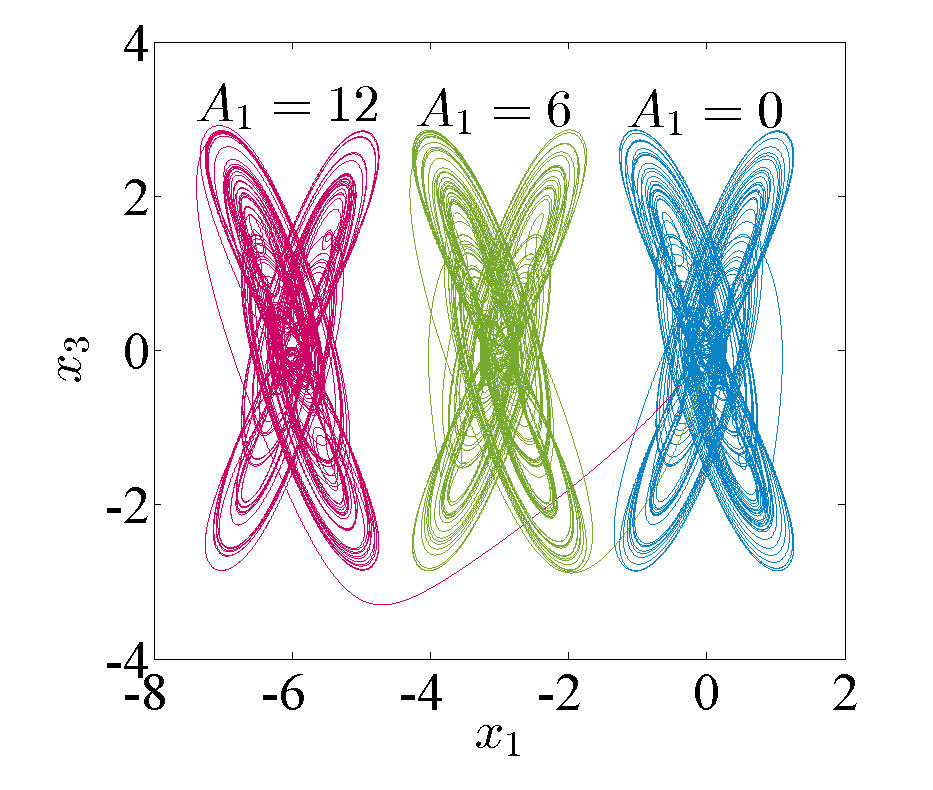}\\
  (c)
 \end{minipage}
 \begin{minipage}{\twofigwidth}
  \centering
  \includegraphics[width=\twofigwidth]{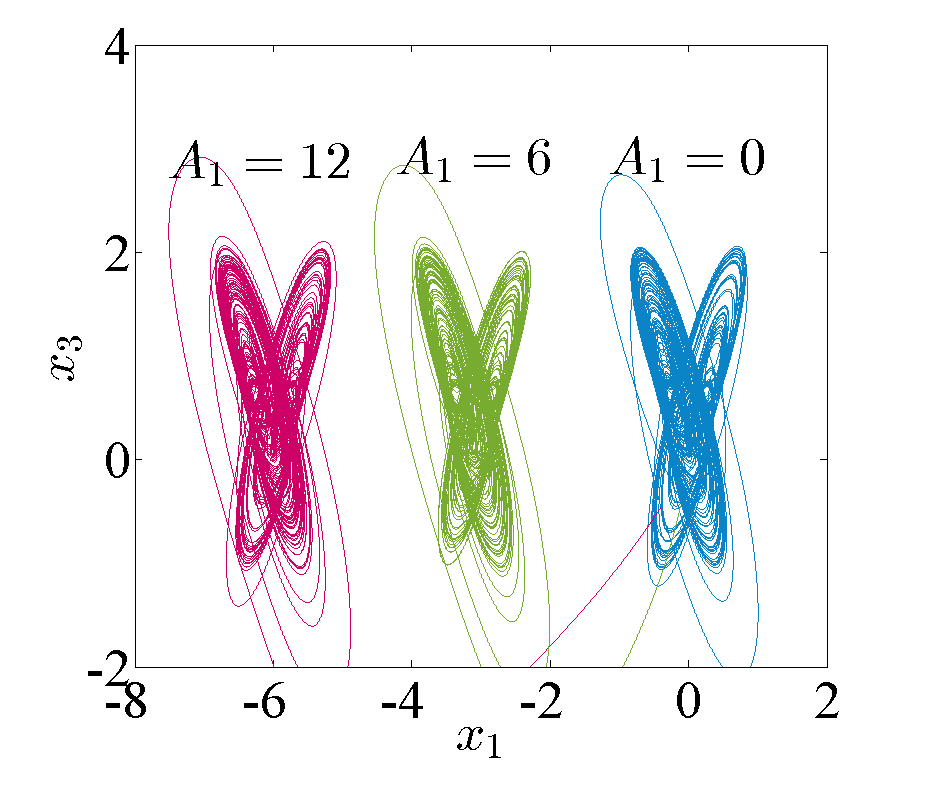}\\
  (d)
 \end{minipage}
 \caption{Phase diagram of AHNN with four WMSs and one SVS:
 (a) eight-scroll attractor; (b) four-scroll attractor; (c) four-scroll attractor with different $A_1$; (d) two-scroll attractor with different $A_1$.}
\label{fig:oneSVSattractor}
\end{figure}

\begin{figure}[!htb]
 \centering
 \begin{minipage}{0.95\twofigwidth}
  \centering
  \includegraphics[width=0.95\twofigwidth]{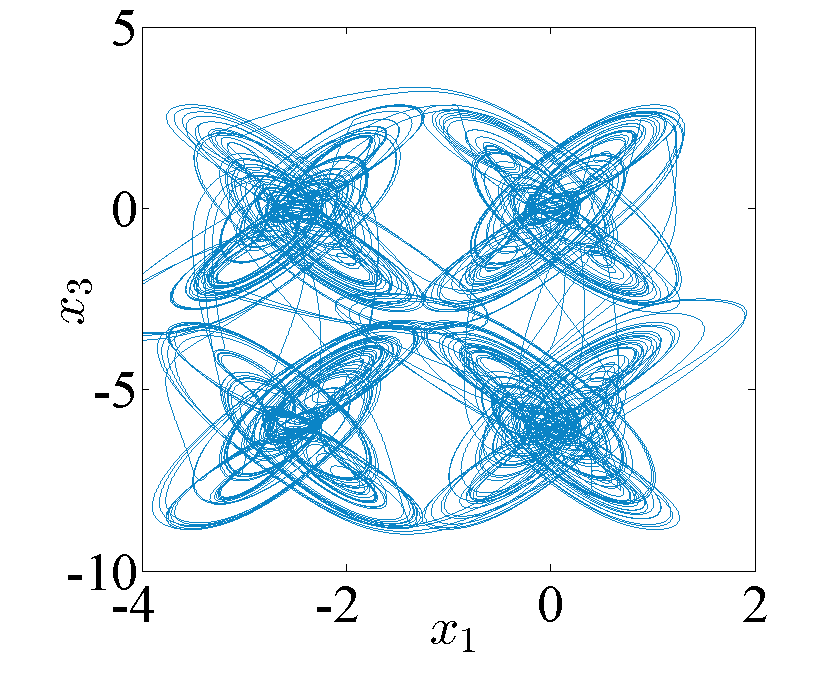}\\
  (a)
 \end{minipage}
 \begin{minipage}{\twofigwidth}
  \centering
  \includegraphics[width=0.95\twofigwidth]{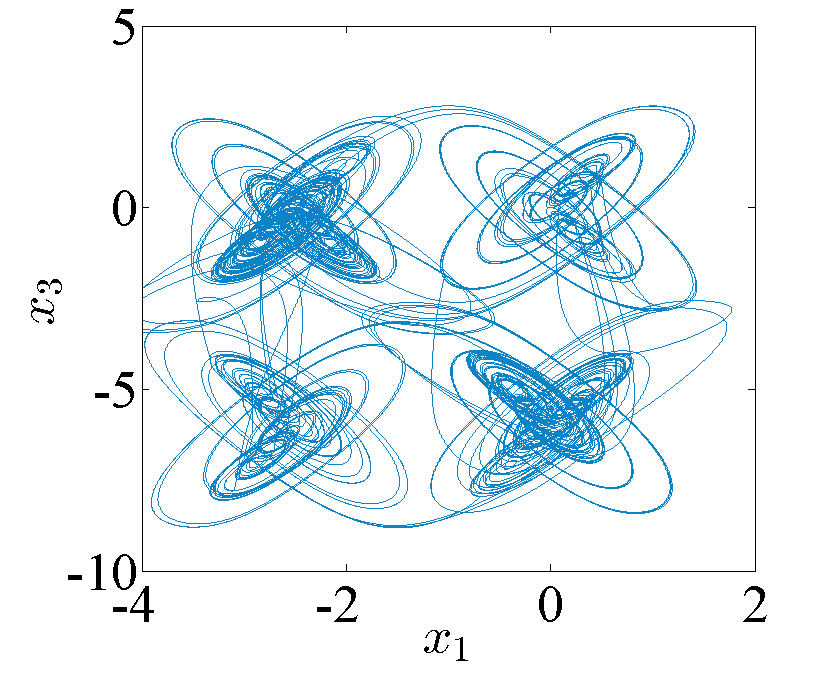}\\
  (b)
 \end{minipage}
 \begin{minipage}{\twofigwidth}
  \centering
  \includegraphics[width=\twofigwidth]{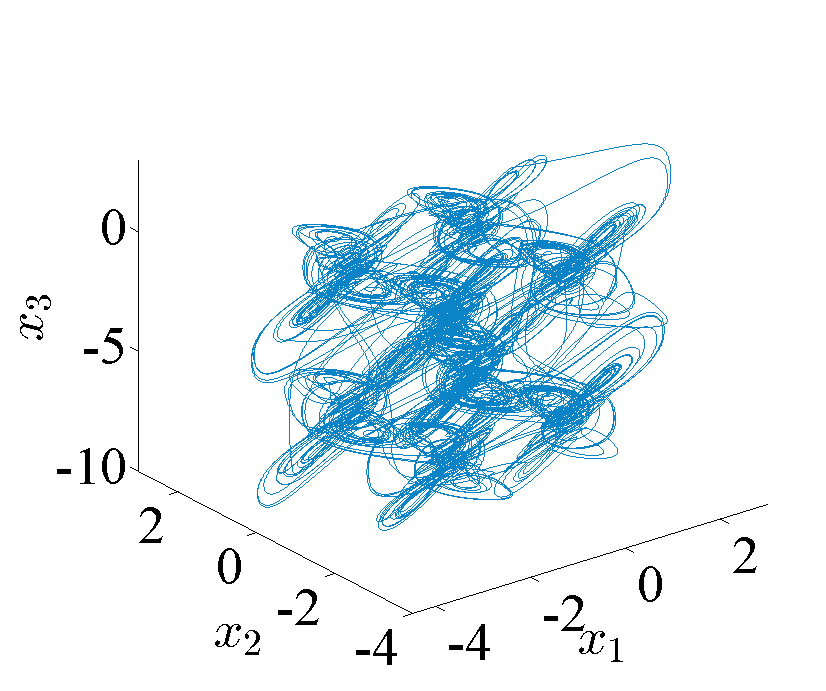}\\
  (c)
 \end{minipage}
 \begin{minipage}{\twofigwidth}
  \centering
  \includegraphics[width=\twofigwidth]{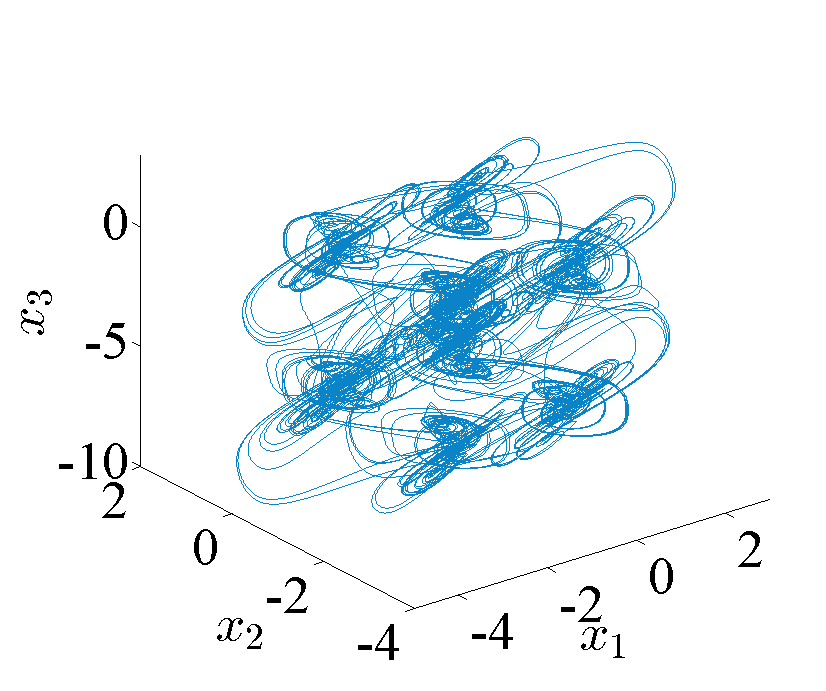}\\
  (d)
 \end{minipage}
\caption{Phase diagram of AHNN with four WMSs and multiple SVSs:
(a) $2\times2$ grid 16-scroll attractor; (b) $2\times2$ grid 9-scroll attractor; (c) $2\times2\times2$ grid 32-scroll attractor; (d) $2\times2\times2$ grid 18-scroll attractor.}
\label{fig:multiSVSattractor}
\end{figure}

\begin{figure}[!htb]
 \centering
 \begin{minipage}{\twofigwidth}
  \centering
  \includegraphics[width=\twofigwidth]{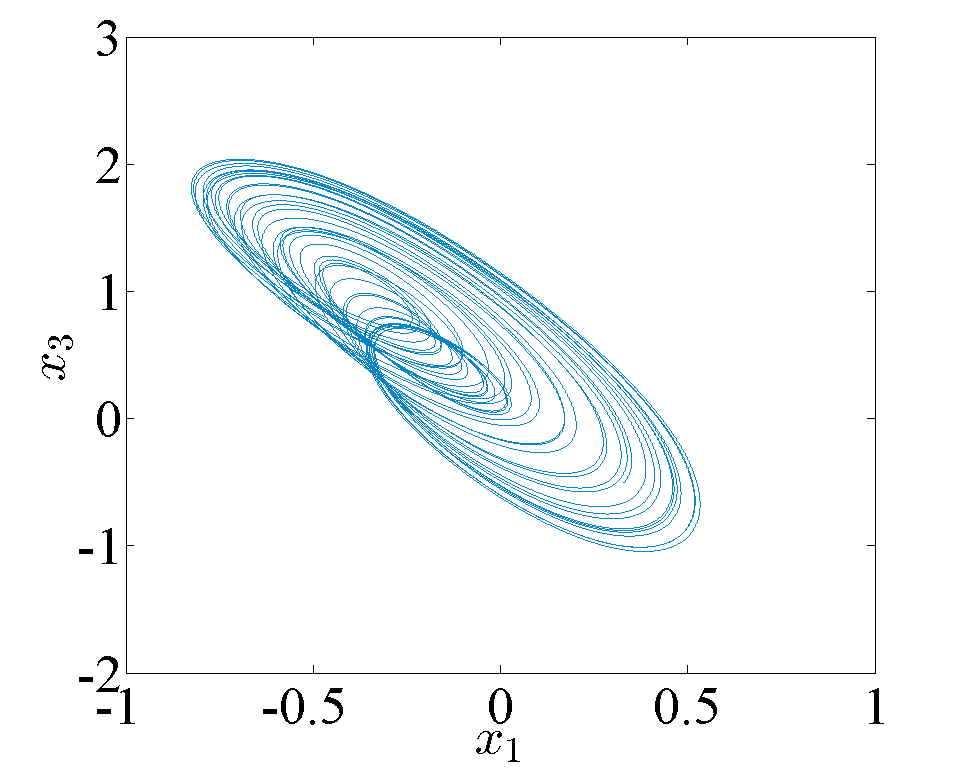}\\
  (a)
 \end{minipage}
 \begin{minipage}{\twofigwidth}
  \centering
  \includegraphics[width=\twofigwidth]{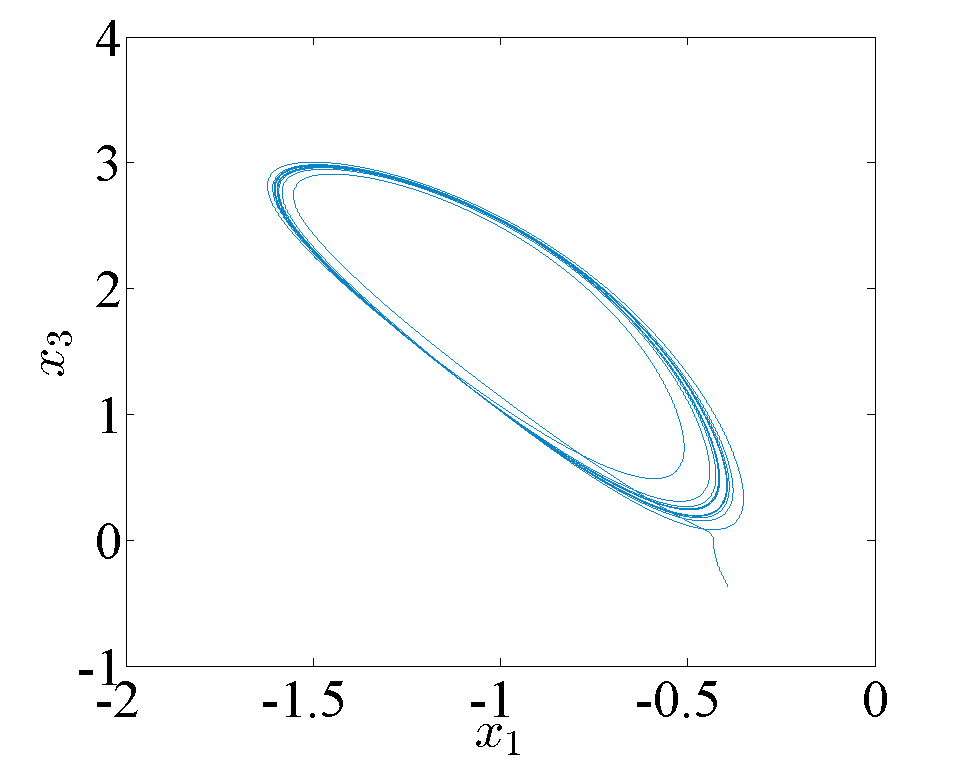}\\
  (b)
 \end{minipage}
 \begin{minipage}{\twofigwidth}
  \centering
  \includegraphics[width=\twofigwidth]{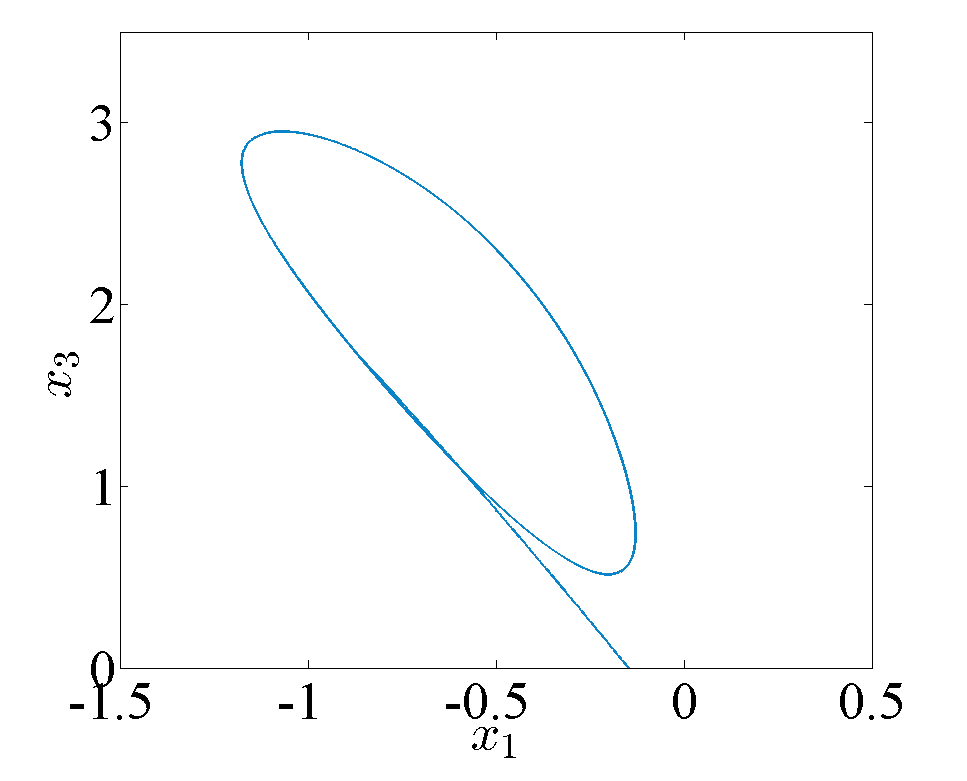}\\
  (c)
 \end{minipage}
 \begin{minipage}{\twofigwidth}
  \centering
  \includegraphics[width=\twofigwidth]{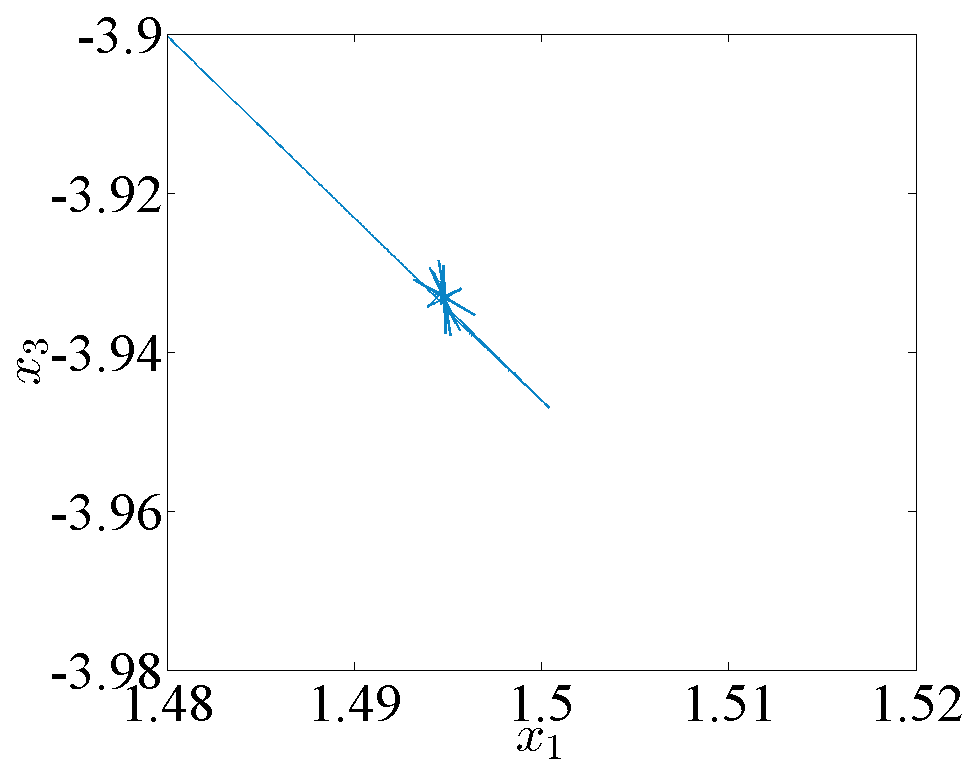}\\
  (d)
 \end{minipage}
\caption{Phase diagram of AHNN with one CS and One SVS: (a) one-scroll attractor when $A'_1=0.1$; (b) quasi-period attractor when $A'_1=0.4$; (c) period attractor when $A'_1=0.85$; (d) point attractor when $A'_1=1$.}
\label{fig:adSVSattractor}
\end{figure}

The AHNN can be categorized into four distinct cases, each defined by the type and number of stimuli applied:
\begin{itemize}
\item Four WMSs:
With $A=1$, $\omega=0.01$, $k=1.15$, and initial conditions $(x_{1}(0), x_{2}(0), x_{3}(0))=(0, 0.1, 0)$, AHNN is observed to generate a four-scroll chaotic attractor, as shown in Fig.~\ref{fig:WMSattractor}(a). When parameter $k$ is adjusted to 1, AHNN demonstrates the ability to create coexisting pairs of two-scroll chaotic attractors, depicted in Fig.~\ref{fig:WMSattractor}(b). The differential effects of the stimuli are further analyzed by presenting time series of $x_i(t) - x'_i(t)$ in Fig.~\ref{fig:WMStime}, where $x_i(t)$ and $x'_i(t)$ denote AHNN solutions with $P(t) = 1$ and $P(t) = -1$, respectively.

\item Four WMSs and one SVS: 
When $ A_1=5$, $\omega_1=0.02$, $A=1$, $\omega=0.01$, $k=1.15$, and initial conditions $(x_{1}(0), x_{2}(0), x_{3}(0))=(0, 0.1, 0)$, AHNN can produce an eight-scroll chaotic attractor, as illustrated in Fig.~\ref{fig:oneSVSattractor}(a). Adjusting $k$ to 1 results in a four-scroll chaotic attractor, shown in Fig.~\ref{fig:oneSVSattractor}(b). The influence of SVS amplitude on the network is explored by plotting chaotic attractors under different $A_1$ in Fig.~\ref{fig:oneSVSattractor}(c) and (d), where the distance between the original and new attractors increases with the amplitude of SVS.

\item Four WMSs and multiple SVSs: 
In this scenario, AHNN exhibits the ability to produce a 2D grid multi-scroll attractor when $A=1, A_1=5, A_3=12$, $\omega=0.01$, $\omega_1=0.02, \omega_3=0.022$, and $k=1.15$, as illustrated in Fig.~\ref{fig:multiSVSattractor}(a). Conversely, when $A_1=A_2=5$, $A_3=12$, $\omega_1=0.02$, $\omega_2=0.021$, $\omega_3=0.022$, AHNN generates a 3D grid multi-scroll attractor, as depicted in Fig.~\ref{fig:multiSVSattractor}(c). The grid attractors comprise different numbers of four-scroll attractors, indicating that each SVS can double the attractor of AHNN and expand the chaotic range in corresponding directions. Simultaneously, $2\times 2$ grid multi-scroll attractor and $2\times 2\times 2$ grid multi-scroll attractor are observed in Fig.~\ref{fig:multiSVSattractor}(b) and (d) when the parameter $k$ is changed to 1.

\item One CS and one SVS:
Using $A'_1$ as CS, AHNN is
\begin{equation}
\centering
\dot{\x}=-
 \begin{bmatrix}
  x_1-A'_1 \\
  x_2 \\
  x_3
 \end{bmatrix}
 +
 W
 \cdot
 \begin{bmatrix}
  \tanh (x_1+P_1(t))\\
  \tanh (x_2)\\
  \tanh (x_3)
 \end{bmatrix}.
 \label{eq:twosvsHNN}
\end{equation}
In this case, AHNN transitions from chaotic behavior to periodic and quasi-periodic states, eventually converging to a fixed point as the amplitude of CS increases. This transformation is illustrated in Fig.~\ref{fig:adSVSattractor}, showcasing the degeneration of chaos.
\end{itemize}

\section{Dynamical characteristics of AHNN}
\label{sec:dynamics}

This section investigates the boundness, equilibrium stability, Lyapunov exponent, and bifurcation of AHNN. 

\subsection{Boundness}

Referring to \cite{Zhang:bound:CNSNS11}, the Lyapunov function of the network~\eqref{eq:HNNstimuli} is
\begin{equation} 
V(\mathbf{x}) = \sum_{i=1}^{3} (x_i + A_i)^2. 
\label{eq:boundedness}
\end{equation}
Its corresponding boundary is
\begin{equation}
L_k = (0.7b - 2.2)^2 + (k + 1.5 + 2b)^2 + (5b + 1)^2, 
\end{equation}
where $b > 1$. The time derivative of $V(\mathbf{x})$ in Eq.~\eqref{eq:boundedness} is
\begin{equation}
\dot{V}(\mathbf{x}) = 2 \sum_{i=1}^{3} (x_i + A_i) \dot{x}_i.
\end{equation}
Given that $\tanh(x_i) < 1$ and $P_i(t) \in \{0, A_i\}$, the inequality can be expressed as
\begin{align}
    \dot{V}(\mathbf{x}) &< 2(x_1 + A_1)(-x_1 - A_1 + 2.2 - 0.7P(t)) + \ldots \nonumber \\
    &\hspace{15pt}+ 2(x_3 + A_3)(-x_3 - A_3 - 5P(t) - 1).  \nonumber \\
    &< 2(x_1 + A_1)(-x_1 - \alpha) + 2(x_2 + A_2)(-x_2 - \beta)  \nonumber \\
    &\hspace{15pt}+ 2(x_3 + A_3)(-x_3 - \gamma)   \nonumber \\
    &< -V(\mathbf{x}) - (x_1 + \alpha)^2 - (x_2 + \beta)^2 - (x_3 + \gamma)^2 + L_k  \nonumber \\
    &< -V(\mathbf{x}) + L_k,
\end{align}
where $\alpha = A_1 + 0.7P(t) - 2.2$, $\beta = A_2 - (k + 1.5 + 2P(t))$, and $\gamma = A_3 + 5P(t) + 1$. When $V(\mathbf{x}) < L_k$, it implies a bounded state with $V(\mathbf{x})$ confined within $L_k$. For $V(\mathbf{x}) > L_k$, defining $U(\mathbf{x}) = V(\mathbf{x}) - L_k$ leads to
\begin{equation}
\dot{U}(\mathbf{x}) + U(\mathbf{x}) < 0.
\end{equation}
Multiplying this inequality by $e^t$ and integrating from $t_0$ to $\tau$ yields
\begin{equation}
e^{\tau} U(\mathbf{x}) < e^{t_0} U(\mathbf{x}(t_0)).
\end{equation}
Substituting $U(\mathbf{x})$ with $V(\mathbf{x}) - L_k$ and $\tau$ with $t$, one can obtain
\begin{equation}
    V(\mathbf{x}) - L_k < (V(\mathbf{x}(t_0)) - L_k) e^{-(t - t_0)}.
\end{equation}
Consequently, $\overline{\lim}_{t \to \infty} V(x_i(t)) < L_k$, indicating that for infinite time, the network's state remains within the boundary $L_k$. Hence, one concludes that
\begin{equation}
    \Omega_k = \left\{\mathbf{x} \, \Big\vert \, \sum_{i=1}^{3} (x_i + A_i)^2 < L_k \right\}
\end{equation}
forms the globally exponentially attractive set of the network~\eqref{eq:HNNstimuli}.

\subsection{Equilibrium stability}

The Jacobian matrix of network~\eqref{eq:OHNN} is 
\begin{equation}
    J = [w_{mn}]_{3\times3} =
    \begin{bmatrix}
        \frac{(6-11h(x_1))}{5} & \frac{(1-6h(x_2))}{5} & \frac{(h(x_3)-1)}{2} \\
        2(h(x_1)-1) & \frac{(1-3h(x_2))}{2} & k(1-h(x_3)) \\
        5(1-h(x_1)) & 0 & h(x_3)-2
    \end{bmatrix},
\end{equation}
Setting $\dot{\mathbf{x}}$ in network~\eqref{eq:OHNN} as zero, one can find
an equilibrium at (0, 0, 0). The characteristic equation at this equilibrium point is
\begin{equation}
    |\lambda I - J_{\text{w}}| = \lambda^3 + 0.3\lambda^2 + 2.1\lambda - 6k + 4.75.
\end{equation}
The eigenvalues at this point are $\lambda_1 = 0.74$, $\lambda_{2, 3} = -0.52 \pm 1.61\iu$ for $k = 1.15$; and $\lambda_1 = 0.50$, $\lambda_{2, 3} = -0.40 \pm 1.53\iu$ for $k = 1$. Therefore, this equilibrium is an unstable saddle focus of index-1, characterized by one positive real root and a pair of conjugate roots with negative real parts.

The Jacobian matrix of network~\eqref{eq:HNNstimuli} with $P_i(t) = 0$ is
\begin{equation}
    J_{\text{w}} =
    \begin{bmatrix}
        w_{11} & P(t)w_{12} & P(t)w_{13} \\
        P(t)w_{21} & w_{22} & w_{23} \\
        P(t)w_{31} & w_{32} & w_{33}
    \end{bmatrix}.
\end{equation}
Setting $\dot{\mathbf{x}}$ to zero in this network, an equilibrium is found at $(0, 0, 0)$. The characteristic equation at this point is
\begin{multline}
    |\lambda I - J_{\text{w}}| = \lambda^3 + 0.3\lambda^2 + (4.9P^2(t)-2.8)\lambda \\
    -6P^2(t)k + 3.55P^2(t) + 1.2.
\end{multline}
The eigenvalues here are identical to those of network~\eqref{eq:OHNN} at its equilibrium since $P^2(t) = 1$. Consequently, the equilibrium of~\eqref{eq:HNNstimuli} with $P_i(t) = 0$ is also an unstable saddle focus of index-1.

When introducing SVSs $P_1(t)$, $P_2(t)$, and $P_3(t)$ to network~\eqref{eq:HNNstimuli}, various equilibria emerge, including $(0, 0, 0)$, $(-A_1, 0, 0)$, $(0, -A_2, 0)$, $(0, 0, -A_3)$, $(-A_1, -A_2, 0)$, $(-A_1, -A_2, -A_3)$, $(-A_1, 0, -A_3)$, and $(0, -A_2, -A_3)$. The eigenvalues of AHNN at the equilibrium $(-A_1, 0, 0)$ are determined as following steps.
The Jacobian matrix of AHNN with $P_2(t)=P_3(t)=0$, and $P(t)=1$ is
\begin{equation}
    J_{\rm s} =
    \begin{bmatrix}
        \frac{(6-11h_1(x_1))}{5} & \frac{(1-6h(x_2))}{5} & \frac{(h(x_3)-1) }{2} \\
        2(h_1(x_1)-1) & \frac{(1-3h(x_2))}{2} & k(1-h(x_3)) \\
        5(1-h_1(x_1)) & 0 & h(x_3)-2
    \end{bmatrix}, 	
    \label{eq:Jacobianws}
\end{equation}
where $h_1(x_1)=\tanh^2(x_1+P_1(t))$. Substituting the equilibrium $(-A_1, 0, 0)$ into Eq.~\eqref{eq:Jacobianws}, the characteristic equation becomes
\begin{multline}
    |\lambda I-J_{\rm s}| = \tanh^2(P_1(t)-x_a)(2.2\lambda^2-1.6\lambda+1.15) \\
    + \lambda^3 + 0.3\lambda^2 + 2.1\lambda - 2.15,  
\end{multline} 
where $x_a$ is the $x$-coordinate value of the equilibrium, defined as
\begin{equation}
    x_a =
    \begin{cases}
        0   & \text{if } P_1(t) = 0; \\
        A_1 & \text{otherwise}.
    \end{cases}
\end{equation}
Since $\tanh^2(P_1(t)-x_a)=0$, the eigenvalues at $(-A_1, 0, 0)$ are equivalent to those at $(0, 0, 0)$. The eigenvalues of network~\eqref{eq:HNNstimuli} with $P_i(t) \neq 0$ and $k=1.15$ at other equilibria are similarly derived.

For $k=1$, network~\eqref{eq:HNNstimuli} exhibits eight equilibria, which can be categorized into two groups. The first group contains (0, 0, 0) and $(-A_1, -A_2, -A_3)$, with corresponding eigenvalues 0.50 and $-0.40 \pm 1.53\iu$, designating these equilibria as unstable saddles of index-1. The second group includes six equilibria: ($0.43, -0.04, 1.2-A_3$), ($0.43, -0.04-A_2, 1.2$), ($0.43, -0.04-A_2, 1.2-A_3$), ($-0.43-A_1, 0.04, -1.2$), ($-0.43-A_1, 0.04, -1.2-A_3$), and ($-0.43-A_1, 0.04-A_2, -1.2$), with eigenvalues -0.72, $0.37 \pm 1.31\iu$. These points are classified as saddle-foci of index-2, characterized by one negative real root and a pair of conjugate roots with a positive real part.

\subsection{Lyapunov exponents}

Lyapunov exponents are pivotal in characterizing the dynamic behavior of chaotic systems. The Lyapunov exponent of a continuous system is
\begin{equation}
\var{LE}_i=\lim_{N \to \infty}\frac{1}{Nt} \sum_{n=1}^N\ln\|{\bf e}_i^{(n)}\|, 
\label{eq:LEs}
\end{equation}
where ${\bf e}_i^{(n)}$ represents a base of the 3-D subspace in the tangent space, $N$ is the iteration number, and $t$ is the calculation time \cite{Shimada:exponents:PTP79}. In this subsection, the impact of stimuli on the Lyapunov exponent of AHNN is analyzed. AHNN with four WMSs exhibits the same Lyapunov exponent for $P(t) \in \{-A, A\}$. AHNN with SVS demonstrates constant Lyapunov exponents. These findings are presented in Property~\ref{prp:lyapunov_wms} and Property~\ref{prp:lyapunov_svs}.

\begin{Property}
The $i$-th Lyapunov exponent of network~\eqref{eq:OHNN} is equivalent to that of network~\eqref{eq:HNNstimuli} with $P_i(t) = 0$.
\label{prp:lyapunov_wms}
\end{Property}
\begin{proof}
Considering network~\eqref{eq:OHNN}, one derives
\begin{equation}
 \centering
 \begin{bmatrix}
  \dot{X} \\
  \dot{x_2} \\
  \dot{x_3}
 \end{bmatrix}
=-
 \begin{bmatrix}
  X \\
  x_2\\
  x_3
 \end{bmatrix}
+\begin{bmatrix}
  2.2 & -1.2 & 0.5 \\
  -2 & 1.5 & k \\
  5 & 0 & -1
 \end{bmatrix}
 \cdot
 \begin{bmatrix}
  \tanh (X)\\
  \tanh (x_2)\\
  \tanh (x_3)
 \end{bmatrix}, 
\label{eq:mHNN}
\end{equation}
where $X = -x_1$. The Jacobian matrix $J'$ of network~\eqref{eq:mHNN} coincides with $J_{\rm w}$ when $P(t) = -1$.

At the first iteration $n=1$, given $(x_{1}(0), x_{2}(0), x_{3}(0)) = (0, 0.1, 0)$ and ${\bf e}_{(i)}^{(1)} = {\bf I}$, the first iterative vectors of network~\eqref{eq:OHNN} are ${\bf e}_{i}^{(2)} = [e_{ij}^{(2)}]$, where $i, j \in \{1, 2, 3\}$. The corresponding vectors of network~\eqref{eq:mHNN} are ${\bf e'}_1^{(2)} = [e_{11}^{(2)}, -e_{12}^{(2)}, -e_{13}^{(2)}]^\mathrm{T}$ and ${\bf e'}_m^{(2)} = [-e_{m1}^{(2)}, e_{m2}^{(2)}, e_{m3}^{(2)}]^\mathrm{T}$ for $m \in \{2, 3\}$.

Assuming that the relation between the networks~\eqref{eq:OHNN} and~\eqref{eq:mHNN} in the $(n-1)$ iteration is the same as that in the first iteration, one has ${\bf e'}_1^{(n-1)} = [e_{11}^{(n-1)}, -e_{12}^{(n-1)}, -e_{13}^{(n-1)}]^\mathrm{T}$ and ${\bf e'}_m^{(n-1)} = [-e_{m1}^{(n-1)}, e_{m2}^{(n-1)}, e_{m3}^{(n-1)}]^\mathrm{T}$. The subsequent iteration then yields $({\bf e'}_1^{(n)}, {\bf e'}_2^{(n)}, {\bf e'}_3^{(n)})^\mathrm{T}$, where
${\bf e'}_1^{(n)} = J'^{(n-1)}{\bf e'}_1^{(n-1)}$,
${\bf e'}_2^{(n)} = J'^{(n-1)}{\bf e'}_2^{(n-1)} - \langle{\bf e'}_1^{(n-1)}J'^{(n-1)}{\bf e'}_2^{(n-1)} \rangle {\bf e'}_1^{(n)} $
and
\begin{multline}
{\bf e'}_3^{(n)} = J'^{(n-1)}{\bf e'}_3^{(n-1)} - \langle {\bf e'}_1^{(n)} J'^{(n-1)}{\bf e'}_3^{(n-1)} \rangle{\bf e'}_1^{(n)} \\
- \langle {\bf e'}_2^{(n)}  J'^{(n-1)}{\bf e'}_3^{(n-1)} \rangle{\bf e'}_2^{(n)}.
\end{multline}
It follows that ${\bf e'}_1^{(n)} = [e_{11}^{(n)}, -e_{12}^{(n)}, -e_{13}^{(n)}]^\mathrm{T}$ and ${\bf e'}_m^{(n)} = [-e_{m1}^{(n)}, e_{m2}^{(n)}, e_{m3}^{(n)}]^\mathrm{T}$.

The Lyapunov exponents of network~\eqref{eq:OHNN} are
\begin{equation}
\var{LE}_{i}^{(n)}=\ln\sqrt{\sum\nolimits_{j=1}^{3}(e_{ij}^{(n)})^2}. 
\end{equation}
As a comparison, the Lyapunov exponents of network~\eqref{eq:mHNN} are
\begin{equation}
\var{LE}_{1}^{(n)} = \ln\sqrt{(e_{11}^{(n)})^2 + \sum\nolimits_{j=2}^{3}(-e_{1j}^{(n)})^2}
\end{equation}
and
\begin{equation}
\var{LE}_{m}^{(n)} = \ln\sqrt{(-e_{m1}^{(n)})^2 + \sum\nolimits_{j=2}^{3}(e_{mj}^{(n)})^2}.
\end{equation}
Consequently, both networks yield identical Lyapunov exponents, affirming the property.
\end{proof}

\begin{Property} 
The $i$-th Lyapunov exponent of network~\eqref{eq:OHNN} is identical to that of network~\eqref{eq:HNNstimuli} when $P_i(t)=0$.
\label{prp:lyapunov_svs}
\end{Property}
\begin{proof}
The network~\eqref{eq:HNNstimuli} with $P(t)=1$ and $P_1(t) \neq 0$ is 
\begin{equation}
 \centering
 \begin{bmatrix}
  \dot{X'} \\
  \dot{x_2} \\
  \dot{x_3}
 \end{bmatrix}
=-
 \begin{bmatrix}
  X' \\
  x_2\\
  x_3
 \end{bmatrix}
+\begin{bmatrix}
  2.2 & -1.2 & 0.5 \\
  2   & 1.5  & k   \\
  -5  & 0    & -1
 \end{bmatrix}
 \cdot
 \begin{bmatrix}
  \tanh (X') \\
  \tanh (x_2)\\
  \tanh (x_3)
 \end{bmatrix}, 
\end{equation}
where $X'= x_1+A_1$.
The corresponding Jacobian matrix is
\begin{equation}
J_{\rm s}=\begin{bmatrix}
\frac{(6-11h(X'))}{5} & \frac{(1-6h(x_2))}{5} & \frac{(h(x_3)-1) }{2}\\
  2(h(X')-1) & \frac{(1-3h(x_2))}{2} & 1.15(1-h(x_3)) \\
  5(1-h(X')) & 0 & h(x_3)-2
 \end{bmatrix}.
\end{equation}
When $(x_{1}(0), x_{2}(0), x_{3}(0))=(0, 0.1, 0)$ and ${\bf e}_{(1, 2, 3)}^{(1)}={\bf I}$, one knows that ${\bf e}_{(1, 2, 3)}^{(n)}$ of AHNN equals to that of the network~\eqref{eq:OHNN}. Thus, the Lyapunov exponent of network~\eqref{eq:HNNstimuli} with $P(t)=1$ and $P_1(t) \neq 0$ equals that of network~\eqref{eq:OHNN}. 
\end{proof}

\begin{figure}[!htb]
\centering
\begin{minipage}{\BigOneImW}
\centering
\includegraphics[width=\BigOneImW]{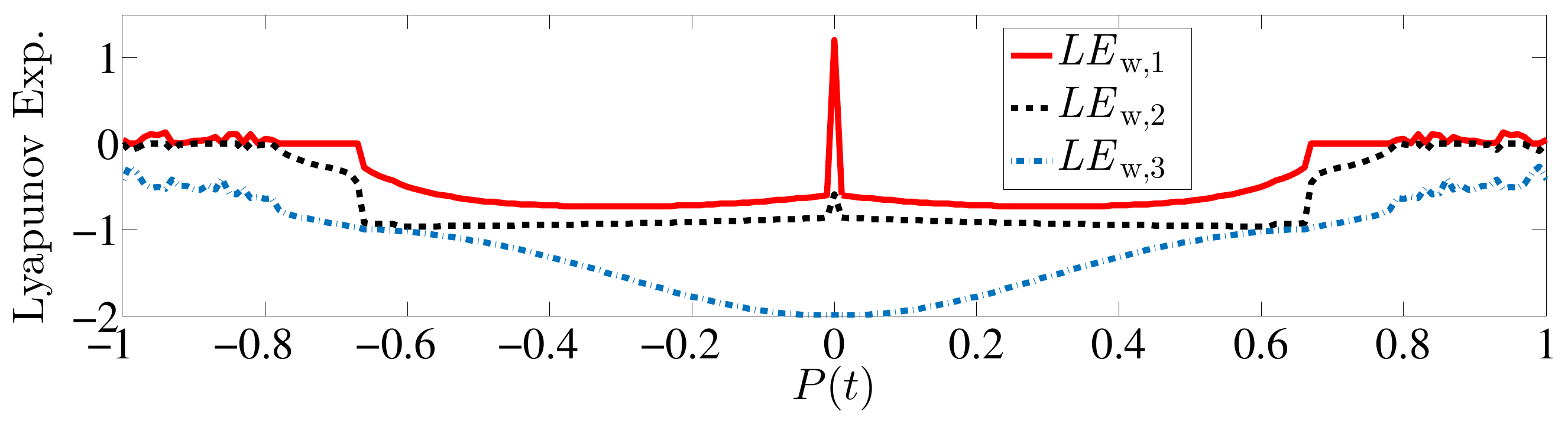}\\
(a)
\end{minipage}\\
\begin{minipage}{\BigOneImW}
\centering
\includegraphics[width=\BigOneImW]{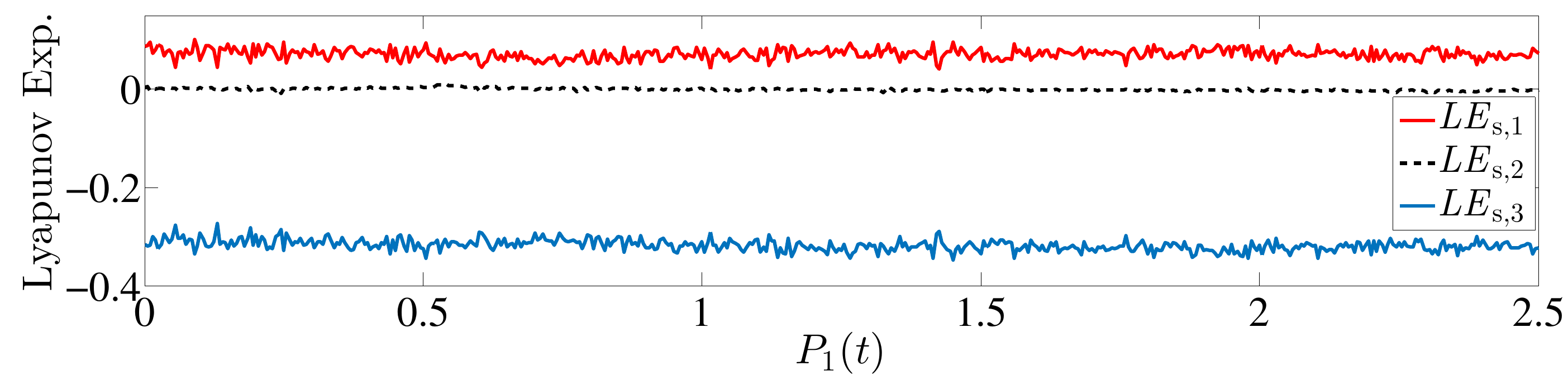}\\
(b)
\end{minipage}
\begin{minipage}{\BigOneImW}
\centering
\includegraphics[width=\BigOneImW]{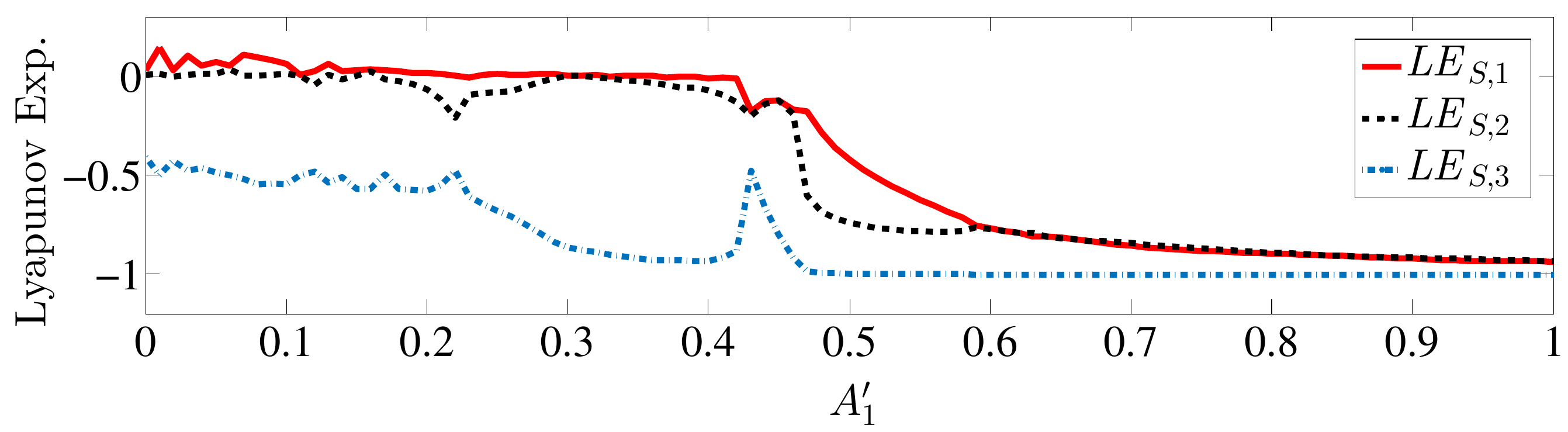}\\
(c)
\end{minipage}
\caption{Lyapunov exponents of network~\eqref{eq:HNNstimuli} adjusted with various stimuli with respect to angular velocity and amplitude: 
(a) four WMSs; (b) four WMSs and one SVS $P_1(t)$; (c) two SVSs.}
\label{fig:Lyexponents}
\end{figure}

Variations in the Lyapunov exponents of AHNN due to WMS $P(t)$ are illustrated in Fig.~\ref{fig:Lyexponents}(a), indicating exponents of 0.066, 0, and $-0.431$ for $P(t)=\pm 1$. The Lyapunov exponents of AHNN with SVS $P_1(t)$, shown in Fig.~\ref{fig:Lyexponents}(b), are approximately 0.08, 0, and $-0.31$. Similarly, the Lyapunov exponents of network~\eqref{eq:twosvsHNN} are displayed in Fig.~\ref{fig:Lyexponents}(c). With increasing parameter $A'_1$, the network loses its positive Lyapunov exponent, leading to the disappearance of its chaotic dynamics.

\subsection{Bifurcation}

\begin{figure}[!htb]
 \centering
 \begin{minipage}{\BigOneImW}
  \centering
  \includegraphics[width=\BigOneImW]{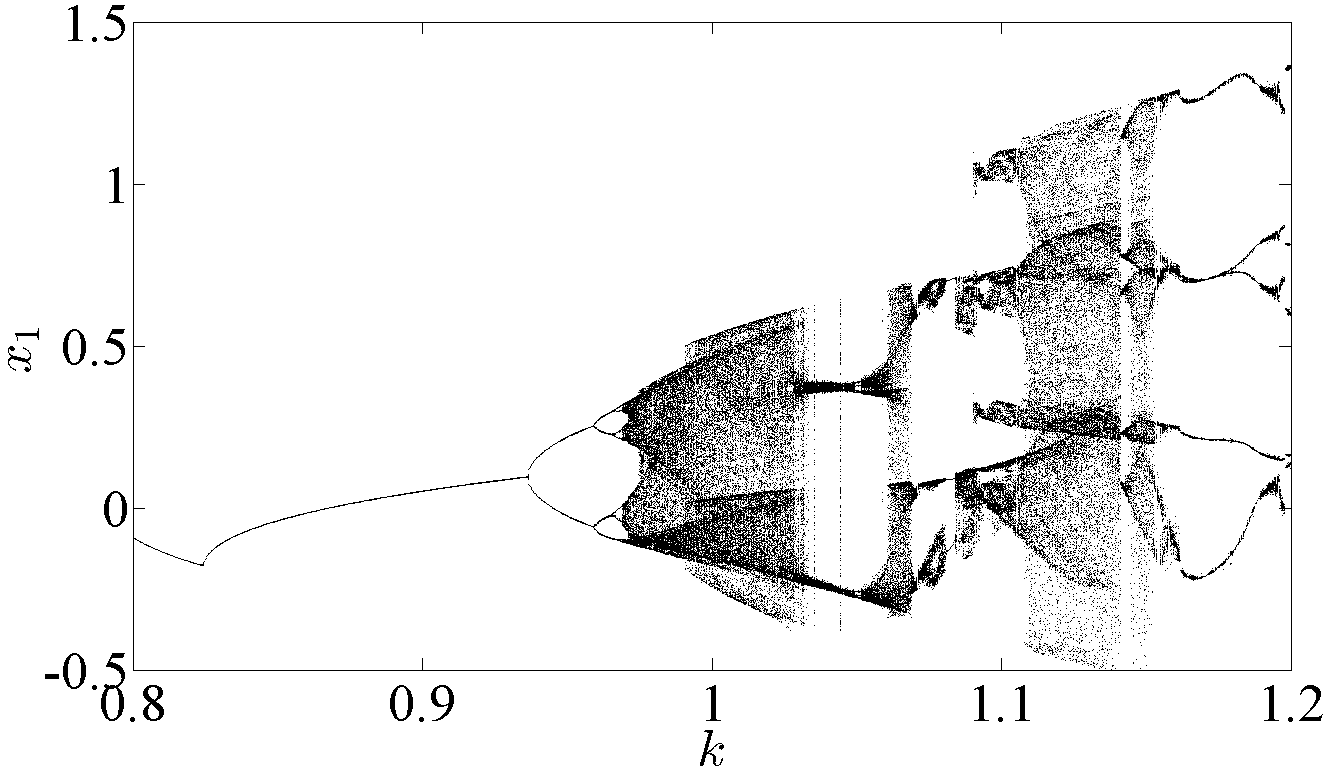}\\
  (a)
 \end{minipage}
 \begin{minipage}{\BigOneImW}
  \centering
  \includegraphics[width=\BigOneImW]{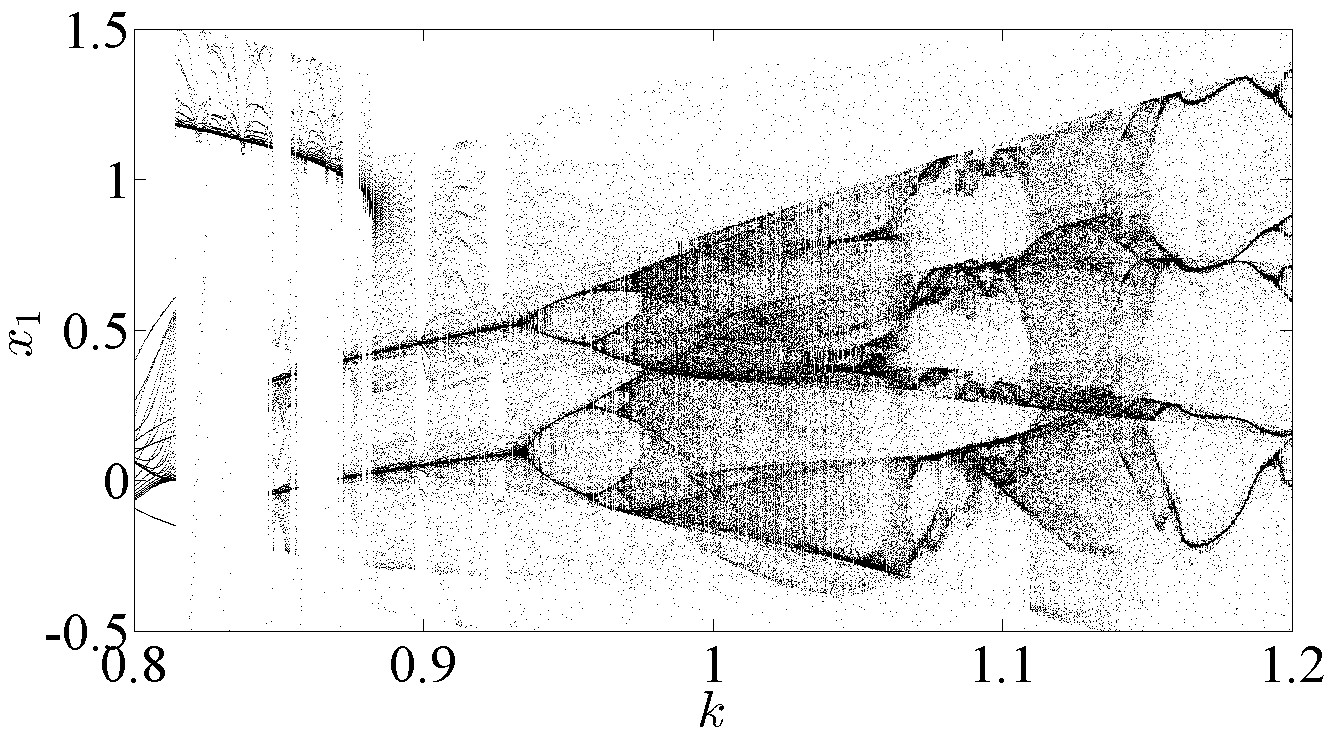}\\
  (b)
 \end{minipage}
  \begin{minipage}{\BigOneImW}
  \centering
  \includegraphics[width=\BigOneImW]{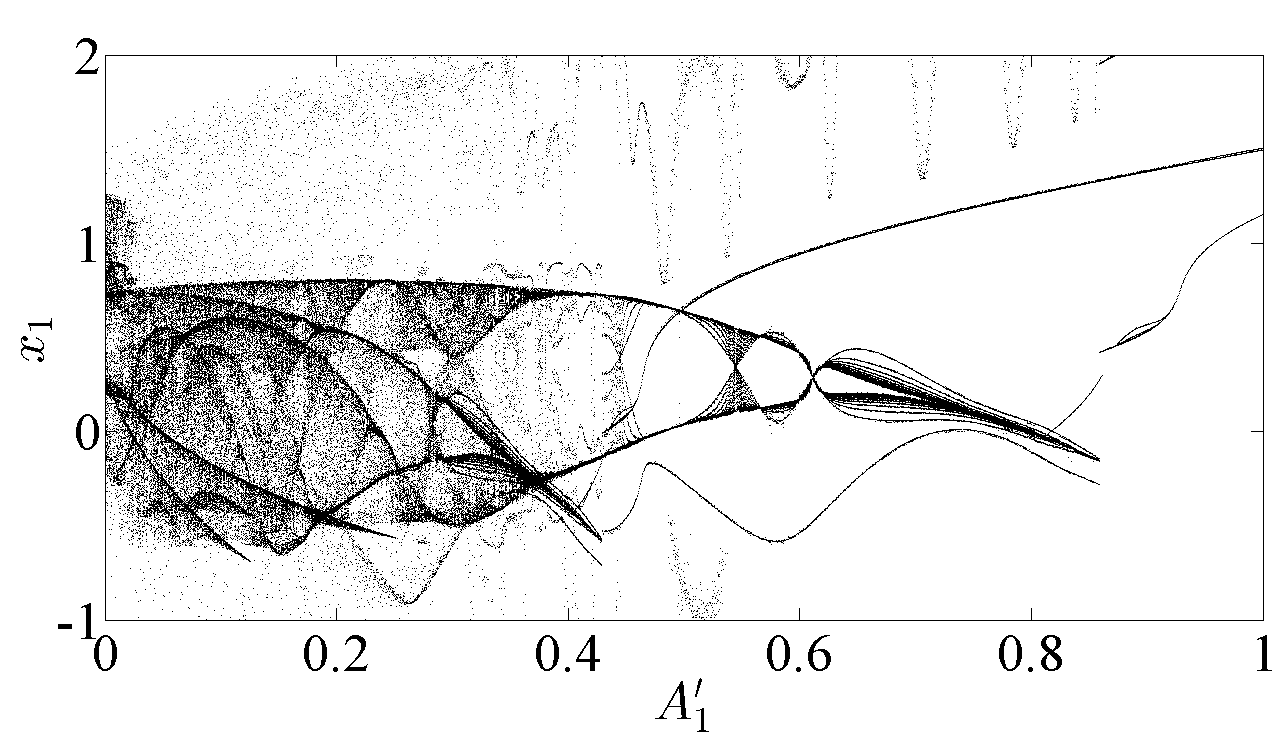}\\
  (c)
 \end{minipage}
\caption{Bifurcation diagram of various models: (a) network~\eqref{eq:OHNN}; (b) network~\eqref{eq:HNNstimuli}; (c) network~\eqref{eq:twosvsHNN}.}
\label{fig:bifurcations}
\end{figure}

Bifurcation diagrams of networks~\eqref{eq:OHNN} and~\eqref{eq:HNNstimuli} are illustrated in Fig.~\ref{fig:bifurcations}(a) and (b). These diagrams reveal that AHNN demonstrates a wider array of chaotic behaviors under external stimuli than in its non-stimulated state. The dynamical behaviors observed in the bifurcation diagrams are consistent with those depicted in the phase portraits of chaotic attractors, confirming the coherence of the results.
Moreover, Fig.~\ref{fig:bifurcations}(c) presents the bifurcation diagram of AHNN with one SVS and one CS, highlighting a notable observation: as the CS's intensity increases, the network's chaotic behavior diminishes. This finding underscores the significant impact of CS's intensity on AHNN's dynamical behavior.

\section{FPGA implementation of AHNN}
\label{sec:FPGA}

Replicating the dynamical behavior of HNN in physical systems is crucial for engineering applications \cite{njitacke:Tabu-analog:TII22, Li:Hopfield:ND22}. Two prevalent methods for HNN implementation are analog circuits and digital platforms. In the analog approach, integrating circuits that simulate differential equations are constructed using discrete components like amplifiers, resistors, transistors, sources, and capacitors \cite{Lin:stimulus:CNSNS2020}. Conversely, digital implementations of HNNs on digital chips employ numerical methods such as Runge-Kutta (RK) or Euler \cite{Wang:FPGAchaos:TCSI2016, Hua:FPGAchaos:TIE2019}. Typically, analog circuits are more resource-intensive and power-consuming due to the extensive use of amplifiers \cite{Lai:Simplechaos:TCSII2020, Wan:analog:CSF23}. Consequently, digital implementations are more practical for replicating HNN's dynamical behavior \cite{Lin:FPGAchaos:TIE2021, Zhang:FPGAchaos:TIE2021}.

\begin{figure}[!htb]
\centering
\includegraphics[width=1.2\BigOneImW]{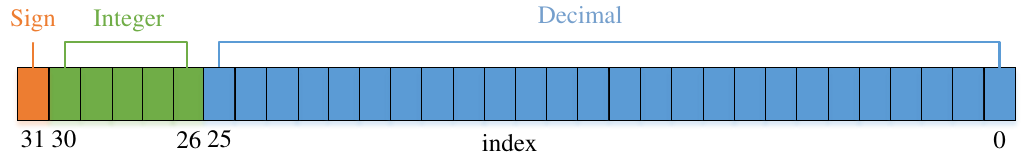}
\caption{The format of a fixed-point number.}
\label{fig:fixed-point}
\end{figure}

In this study, the dynamical behavior of AHNN is implemented on an FPGA platform using chip ``Xilinx xq7z020clg400". The implementation involves the RK-4 numerical method and 32-bit signed fixed-point numbers with a 1-bit sign, 5-bit integer, and 26-bit decimal representation, as shown in Fig.~\ref{fig:fixed-point}. For example, the decimal numbers ``0.012" and ``$-0.125$" are converted to hexadecimal formats ``$000\mathrm{C} 49\mathrm{BA}$" and ``$\mathrm{FF}80 0000$" on the FPGA, respectively.

\begin{figure}[!htb]
 \centering
 \begin{minipage}{\BigOneImW}
  \centering
  \includegraphics[width=\BigOneImW]{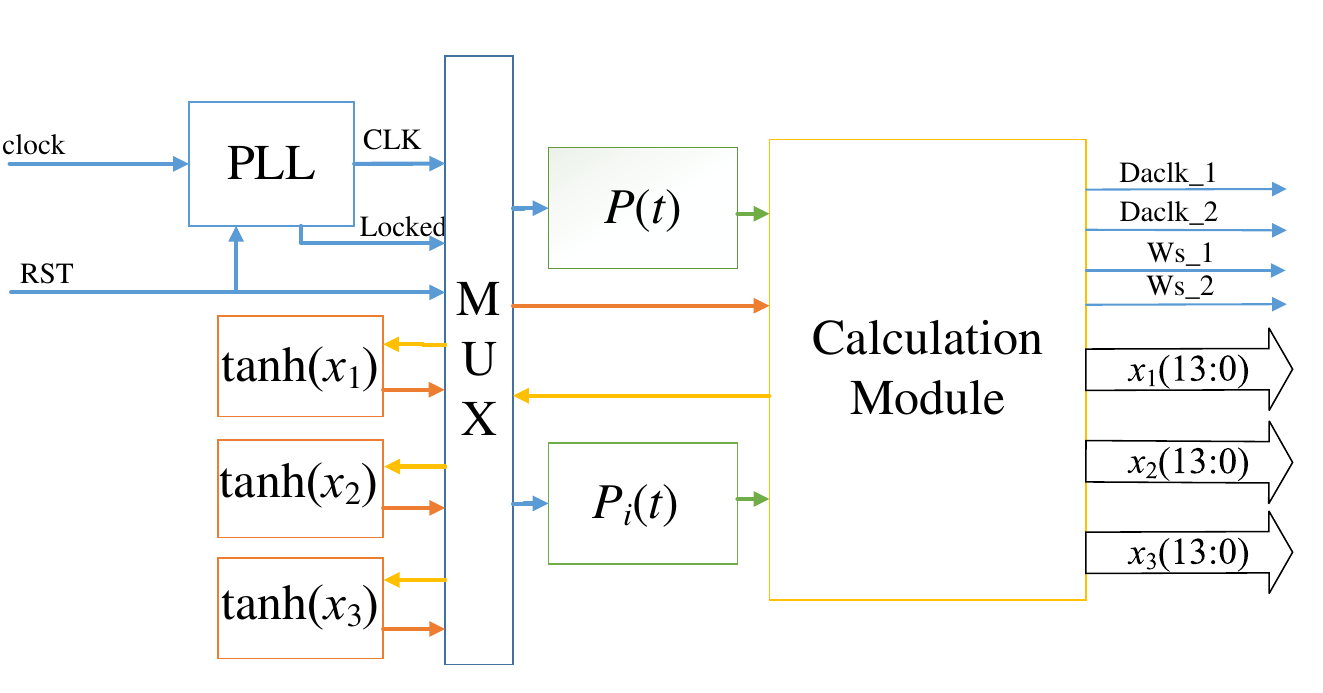}\\
  (a)
 \end{minipage}
 \begin{minipage}{\BigOneImW}
  \centering
  \includegraphics[width=\BigOneImW]{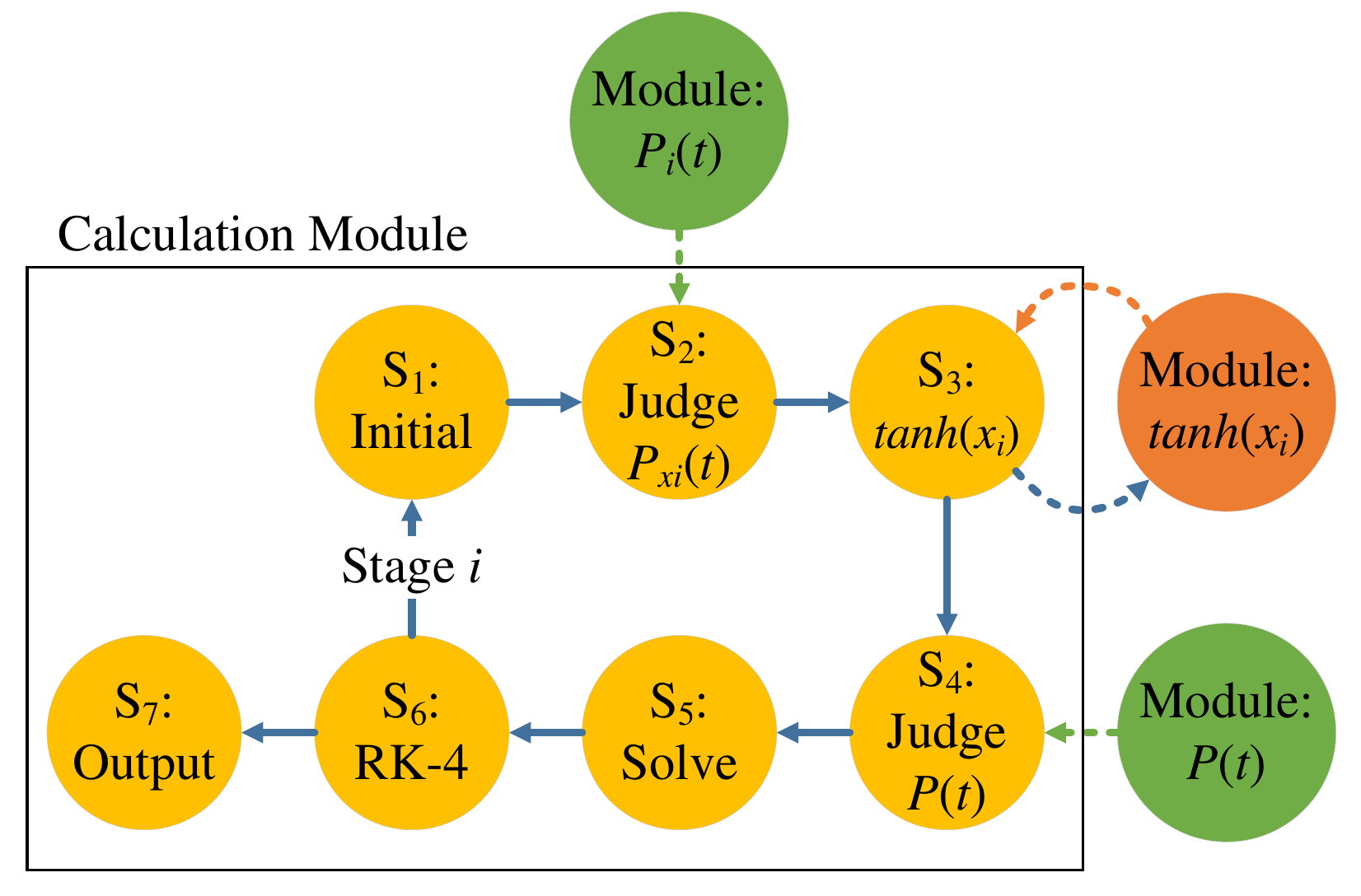}\\
  (b)
 \end{minipage}
 \begin{minipage}{\BigOneImW}
  \centering
  \includegraphics[width=\BigOneImW]{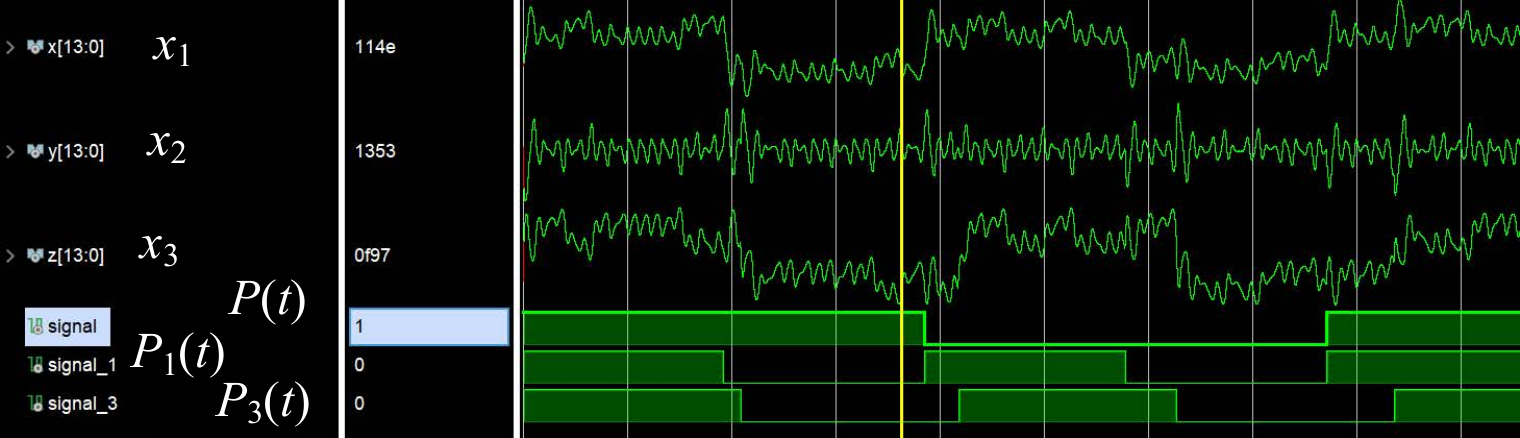}\\
  (c)
 \end{minipage}
\caption{The implementation of AHNN on FPGA platform: (a) the top-level module diagram; (b) state machine flowchart; (c) functional simulation.}
\label{fig:FPGA}
\end{figure}

The top-level module of the implementation features two inputs (drive clock and reset signal) and seven outputs (four digital-to-analog converter drive signals and three state variables). It consists of eight submodules: three hyperbolic tangent function generators, two time-variant stimuli generators, a Phase-Locked Loop (PLL), a multiplexer (MUX), and a calculation module, as detailed in Fig.~\ref{fig:FPGA}(a).

Among the submodules, the hyperbolic tangent function generator is vital for implementing AHNN on the FPGA due to the complexities of directly encoding it in hardware language. To address this, approaches such as Taylor series expansion \cite{Rajagopal:implementation:EPJST2019}, piecewise linear approximation \cite{Tlelo-Cuautle:implementation:2020}, and nonlinear substitution \cite{Zhu:implementation:CW2021} are often utilized. In this work, the Taylor series expansion is employed for its superior accuracy in approximating the hyperbolic tangent function.

\begin{figure}[!htb]
\begin{minipage}{\BigOneImW}
\centering
\includegraphics[width=\BigOneImW]{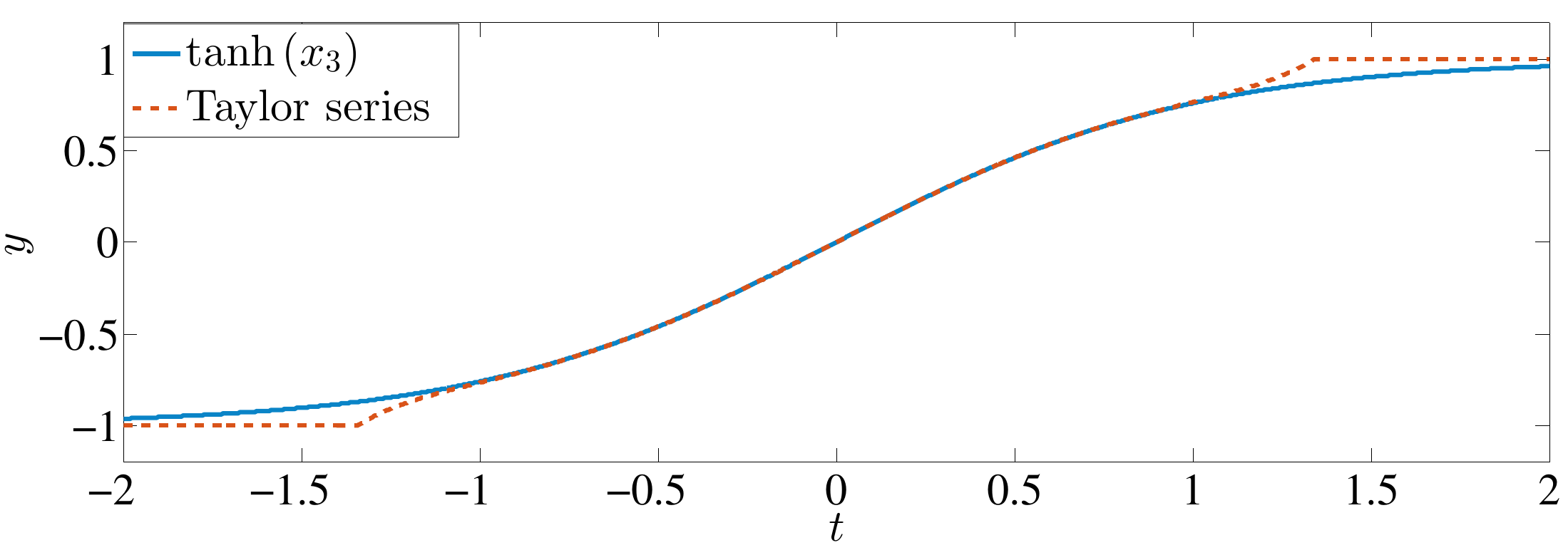}\\
(a)
 \end{minipage}
\begin{minipage}{\BigOneImW}
\centering
\includegraphics[width=\BigOneImW]{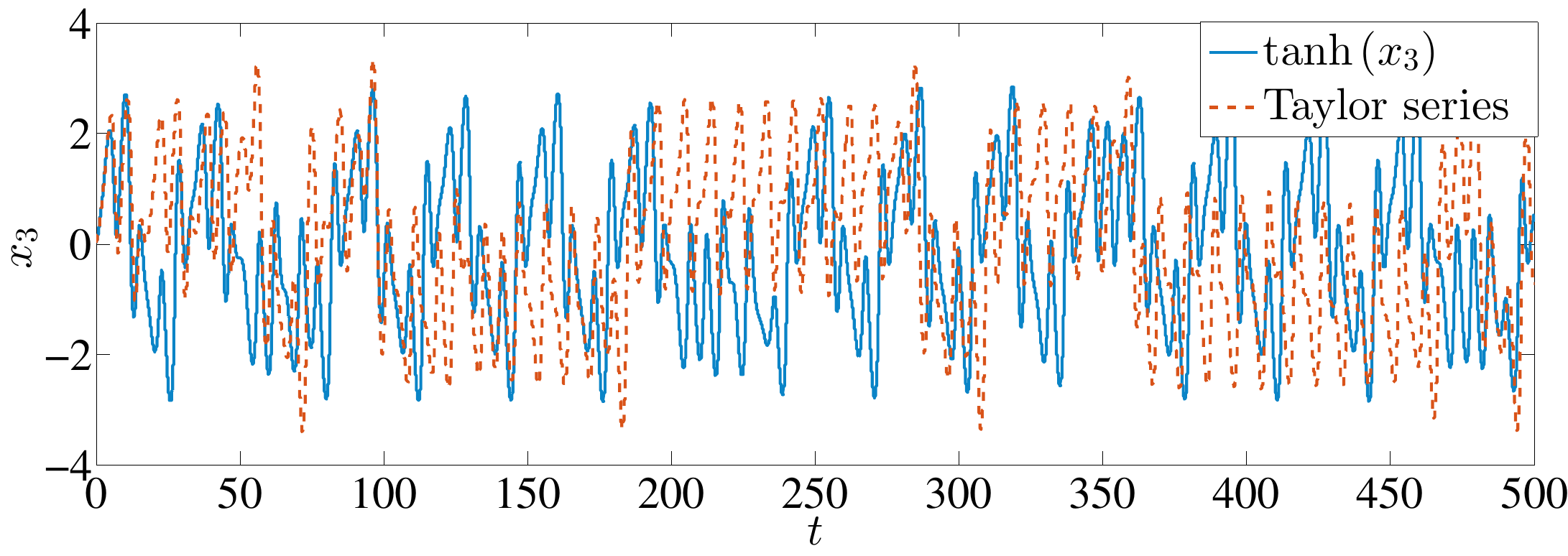}\\
(b)
\end{minipage}
\centering
\caption{Comparison chart between $\tanh(t)$ and its Taylor approximation:(a) fitting effect (b) time serial of AHNN with these functions.}
\label{fig:compare}
\end{figure}

Diverging from \cite{Rajagopal:implementation:EPJST2019}, $\tanh(x)$ is represented by a piece-wise Taylor series
\begin{equation}
h(x)=
\begin{cases}
 -1  & \mbox{if }x\leq -a; \\
x-\frac{x^3}{3}+\frac{2x^5}{15}-\frac{17x^7}{315} +\frac{62x^9}{2835} 
     & \mbox{if }\left| x \right|<a; \\
  1  & \mbox{if } x\ge a, 
\end{cases} 
\label{eq:Taylor}
\end{equation}
where $a$ is a positive real number and $h(x) \leq 1$. The optimal value of $a$ is determined by minimizing the fitting error $\delta(a)$, calculated as
\begin{IEEEeqnarray}{rCl}
 \delta(a) & = & \int_{-\infty}^{+\infty} | h(x)-\tanh(x)|\, dx \nonumber\\
 & = & 2\left|\int_0^a x-\frac{x^3}{3}+\frac{2x^5}{15}-\frac{17x^7}{315}+\frac{62x^9}{2835}-\tanh(x)\, dx\right| \nonumber\\
 &   & \;\;\;+ 2\int_a^{+\infty} 1-\tanh(x)\, dx. 
\end{IEEEeqnarray}
The minimum of $\delta(a)$ is achieved when $a\approx 1.34$. Figure~\ref{fig:compare}(a) shows a comparison chart between the hyperbolic tangent function and its Taylor series approximation with $a=1.34$. Figure~\ref{fig:compare}(b) examines the dynamic degeneration of the network, comparing the AHNN with the actual hyperbolic tangent function and its approximation. The fitting error leads to different sequences after several iterations. While adding higher-order terms to the Taylor series enhances precision, it also increases computation time and resource consumption.

The functions of other submodules are also elaborated. The PLL generates a 50 MHz clock and a ``locked'' control signal to orchestrate the entire system. The time-variant stimulus generator comprises several registers and counters. Upon reaching its maximum value, the counter triggers a change in the number stored in the register, which is then output as the current stimulus. Subsequently, the counter resets to zero and increments by one upon completing an iteration. The core component of the calculation module is a state machine \cite{Wang:FPGA:TCSII2019}, which operates as follows:
\begin{itemize}[itemindent=0em]
\item \textit{State 1}: 
Load the initial conditions $x_1(i)$, $x_2(i)$, $x_3(i)$, with the starting conditions set as (0, 0.1, 0) for the first iteration.

\item \textit{State 2}: 
Compute the state variable using
    \begin{equation}
    x_c(i)=
    \begin{cases}
    x_1(i)        & \text{if }  P_1(t)=0;\\
    x_1(i)+A_1    & \text{otherwise}.
    \end{cases}
    \end{equation}
    Variables $y_c$ and $z_c$ are derived similarly.

\item \textit{State 3}: 
Transmit the current state variables $x_c$, $y_c$, and $z_c$ to the hyperbolic tangent function generator, yielding $\tanh{(x_c)}$, $\tanh{(y_c)}$, and $\tanh{(z_c)}$.

\item\textit{State 4}:  
Define
\begin{equation}
g_x=
\begin{cases}
 -x_c-1.2\tanh{(y_c)}+0.5\tanh{(z_c)}  & \text{if }  P(t) = 1;\\
 -x_c+1.2\tanh{(y_c)}-0.5\tanh{(z_c)}  & \text{otherwise}.
\end{cases}
\end{equation}
Similarly, define $g_y = -y_c\pm 2\tanh{(x_c)}$ and $g_z = -z_c\mp 5\tanh{(x_c)}$.

\item \textit{State 5}:  
Solve
    \begin{equation}
    \left\{
    \begin{aligned}
        G_x &=g_x+2.2\tanh{(x_c)}\\
        G_y &=g_y+1.5\tanh{(y_c)}+k\tanh{(z_c)}\\
        G_z &=g_z-\tanh{(z_c)}\\
    \end{aligned}
    \right.
    \end{equation}
    to obtain the result.

\item \textit{State 6}:  
Update the state variable $x_1(i+1)=x_1(i)+\frac{h}{6}(k_1+2k_2+2k_3+k_4)$, where $h$ is the time interval. For instance, in stage one, the state variable becomes $x_1(i)=x_c+\frac{hG_x}{6}$, and the state machine resets to \textit{State 1}.

\item \textit{State 7}:  
Conclude the iteration and output the results $x_i(i+1)$.
\end{itemize}

The procedural flowchart is depicted in Fig.~\ref{fig:FPGA}(b). The MUX, a network of switches and wires, is responsible for data selection and transmission. The connections between MUX and other submodules facilitate the transfer of 32-bit data and control signals, while the links between the time-variant stimulus generators and the calculation module assign the current values of $P(t)$ and $P_i(t)$.

\begin{figure}[!htb]
 \centering
 \begin{minipage}{0.83\twofigwidth}
  \centering
  \includegraphics[width=0.83\twofigwidth]{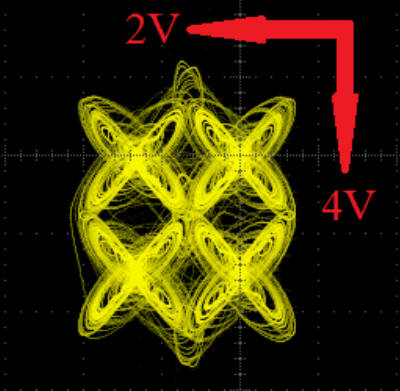}\\
  (a)
 \end{minipage}
 \begin{minipage}{0.8\twofigwidth}
  \centering
  \includegraphics[width=0.8\twofigwidth]{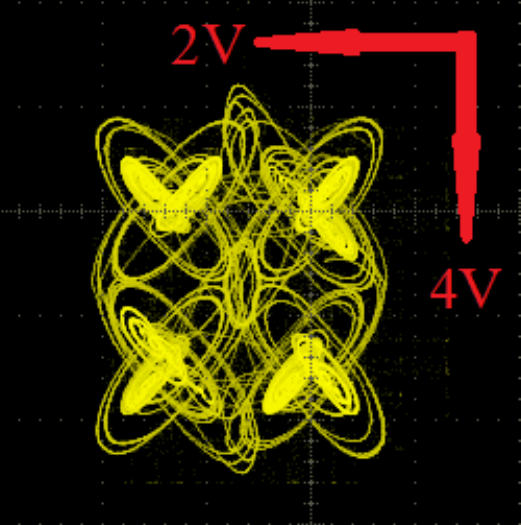}\\
  (b)
\end{minipage}
\caption{Phase diagram of AHNN with four WMSs and multiple SVSs on FPGA platform: (a) $k=1.15$; (b) $k=1$.}
\label{fig:FPGAoutput}
\end{figure}

\begin{table}[!htb]
\renewcommand{\arraystretch}{1.3}
 \caption{Resources estimation of implementation.}
 \centering
\resizebox{\columnwidth}{!}{
  \begin{tabular}{c c c c}  
   \hline\\[-3mm]
   \multicolumn{1}{l}{Resources} & \multicolumn{1}{c}{Utilization} & \multicolumn{1}{c}{Available} & \multicolumn{1}{c}{\pbox{20cm}{Utilization \%}} \\[1.6ex] \hline
   LUT  &  27350  & 53200   & 51.41\%  \\
   FF   &  3362   & 106400  & 3.16\%  \\
   DSP  &  126    & 220     & 57.27\%  \\
   I/O  &  48     & 125     & 38.4\%  \\
   BUFG &  2      & 32      & 6.25\%  \\
   MMCM &  1      & 4       & 25\%  \\
[1.4ex]
   \hline
  \end{tabular}
 }
\label{tab:FPGA}
\end{table}
For illustrative purposes, AHNN with four WMSs and multiple SVSs is used as an example, with its functional simulation shown in Fig.~\ref{fig:FPGA}(c). Utilizing Vivado software's ``synthesis'' and ``implementation'' steps, resource utilization and power consumption are estimated as outlined in Table~\ref{tab:FPGA}. Following the ``generate bitstream'' and ``program'' procedures, the AHNN project is downloaded onto the FPGA chip. The FPGA outputs are then converted to analog signals via a 14-bit digital-to-analog converter AN9767, with experimental results observed on an oscilloscope, as depicted in Fig.~\ref{fig:FPGAoutput}. To assess the implementation's effectiveness, statistical tests are conducted on the time series of AHNN in both simulation and implementation, yielding mean and mean square values of $-0.0045$, 0.3693, 0.0054, and 0.3221.

\section{Application in secure image communication}
\label{sec:algorithm}

\begin{figure*}[!htb]
\centering
\includegraphics[width=2\BigOneImW]{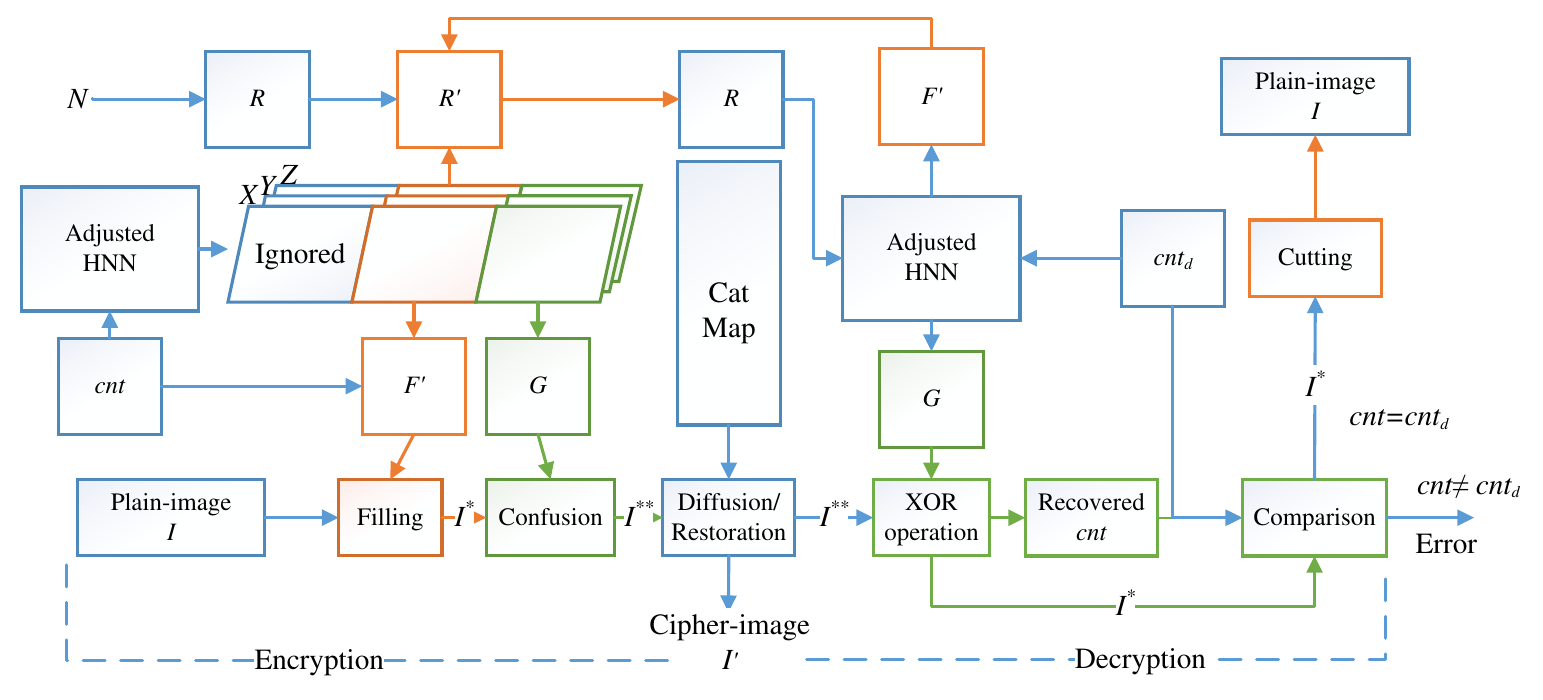}
\caption{The overall framework of the proposed scheme.}
\label{fig:image_algo}
\end{figure*}

\subsection{Encryption scheme}

Due to its high sensitivity to initial conditions, excellent pseudo-randomness, and the long-term unpredictability of orbits, chaos plays a significant role in certain cryptography operations \cite{ZHU:poly:MCS22}. Consequently, various chaotic systems are utilized in designing image encryption schemes \cite{Li:encryption:TNNLS2021, Lakshmanan:secure:TNNLS2018, cqli:block:JISA20, Sarosh:image:TII22}. An image encryption scheme leveraging the advantages of AHNN is proposed, with its architecture depicted in Fig.~\ref{fig:image_algo}. In this scheme, the plain image is represented by $\bm{I}=\{I(i, j)\}^{M, N}_{i=0, j=0}$, where $M$ and $N$ ($M > N$) are the sizes of the image, and the cipher image is denoted by $\bm{I}'=\{I'(i, j)\}^{M, M}_{i=0, j=0}$. The encryption process encompasses the following steps:
\begin{itemize}[leftmargin=*]
\item \textit{The secret key}: This involves the parameters $A_i$, $\omega$, $\omega_i$, and the initial condition $(x_1(0), x_2(0), x_3(0))$ of network~\eqref{eq:HNNstimuli}, along with the parameters of the Cat map:         
\begin{equation}
\left\{
\begin{aligned}
x'_n & =(ax_n+by_n) \bmod N, \\
y'_n & =(cx_n+dy_n+ef(x'_n)) \bmod N,
\end{aligned}
\right.
\label{eq:catmap}
\end{equation}
where $x_n$ and $y_n$ represent the current pixel positions; $x'_n$ and $y'_n$ are the positions of the diffused pixel, with $f(x'_n)=(x'_n)^2+1$, and $a, b, c, d, e$ are real numbers.

\item \textit{Initialization}:
\begin{itemize}[leftmargin=*]
\item \textit{Step 1}:
    A counter $\var{cnt} \in \{0, 1, \cdots, 2^{12}-1\}$ is defined. For each plain image encryption, $\var{cnt}$ increments by one, resetting to zero upon reaching $2^{12}-1$. A 16-bit register $R$ is utilized to store $N$ simultaneously.

\item \textit{Step 2}:
    The secret keys $A_1, A_2, A_3$ are updated as follows:
    \begin{equation}
    \left\{
    \begin{aligned}
    A''_1 &=A_1+\var{cnt}\bmod16 \times 2^{-3}, \\
    A''_2 &=A_2+(\var{cnt} >> 4)\bmod16 \times 2^{-2}, \\
    A''_3 &=A_3+(\var{cnt} >> 8)\bmod16 \times 2^{-3}, 
    \end{aligned}
    \right.
    \end{equation}
    where ``$>>$" signifies the bitwise right shift operation. If $A''_i > \max(A_i)$, the method is adjusted to
    \begin{equation}
    \left\{
    \begin{aligned}
    A''_1 &=A_1-\var{cnt}\bmod16 \times 2^{-3}, \\
    A''_2 &=A_2-(\var{cnt} >> 4)\bmod16 \times 2^{-2}, \\
    A''_3 &=A_3-(\var{cnt} >> 8)\bmod16 \times 2^{-3}. 
    \end{aligned}
    \right.
    \end{equation}

\item \textit{Step 3}:
    Network~\eqref{eq:HNNstimuli} is iterated $g(\var{cnt}, T)+2M^2-M \times N -2$ times using the updated keys, where
    \begin{equation}
    g(\var{cnt}, T)=
    \begin{cases}
    1000               & \mbox {if } \var{cnt} \times T<1000;\\
    \var{cnt} \times T & \mbox {otherwise},
    \end{cases}
    \end{equation}
    and $T\in\{0, 1, \cdots, 5\}$.
    The first $\var{cnt} \times T$ outputs are disregarded, and the remaining outputs are denoted as $\bm{X}=\{X(i)\}_{i=0}^{M(2M-N)-2}, \bm{Y}=\{Y(i)\}_{i=0}^{M(2M-N)-2}$, and $\bm{Z}=\{Z(i)\}_{i=0}^{M(2M-N)-2}$, each being a 32-bit fixed-point number.

\item \textit{Step 4}:
    The sequences $\bm{X}, \bm{Y}$, and $\bm{Z}$ are truncated to 8 bits for image processing. For instance, $X(i)[7:0]$ indicates the seventh to the lowest bit of the $i$-th number in sequence $\bm{X}$. The encryption result of $R$ is
    \begin{equation}
    R'=R \oplus \langle X(0)[7:0], Z(0)[7:0]\rangle,
    \end{equation}
    where $\langle \cdot \rangle$ represents the splicing operation.

\item \textit{Step 5}:
    An 8-bit sequence $\bm{F}=\{F(i)\}_{i=0}^{M(2M-N)-2}$ is obtained as
    \begin{equation}
    F(i)=
    \begin{cases}
    X(i)[7:0]  & \mbox {if } Y(i)[1:0]=00;\\
    Z(i)[7:0]  & \mbox {if } Y(i)[1:0]=01;\\
    Y(i)[9:2]  & \mbox {if } Y(i)[1:0]=10;\\
    Y(i)[15:8] & \mbox {otherwise}.
    \end{cases}
    \end{equation}
    Subsequently, two 8-bit elements are added at the end of this sequence, forming $\bm{F}'=\{F'(i)\}_{i=0}^{M(M-N)}$. These elements are $\langle\var{cnt}[11:8], 0000\rangle$ and $\var{cnt}[7:0]$. For instance, when $\var{cnt}=582$, the elements are 32 and 70.

\item \textit{Step 6}:
The sequence $\bm{G}=\{G(i)\}_{i=0}^{M\times M}$ is generated using the method outlined in Table~\ref{tab:RNG}.
\end{itemize}

\item \textit{The encryption procedure}:
  \begin{itemize}[leftmargin=*]
            \item \textit{Step 1}:
            Populate $\bm{I}$ using the sequence $\bm{F}'$ in raster order to $\bm{I}^*=\{I^*(i, j)\}^{M, M}_{i=0, j=0}$.
            
            \item \textit{Step 2}:
            Perform confusion by XOR operation on $\bm{I}^*$ to obtain output $\bm{I}^{**}=\{I^{**}(i, j)\}^{M, M}_{i=0, j=0}$, defined as
            \begin{equation}
            I^{**}(i)=G(i) \oplus I^*(i).
            \end{equation}
            
            \item \textit{Step 3}:
            Execute a diffusion operation on $\bm{I}^{**}$ using Eq.~\eqref{eq:catmap}.
            
            \item \textit{Step 4}:
            Repeat \textit{Step 3} for $4+\sum\nolimits_{i=0}^3\var{cnt}[i]2^i$ iterations. For example, when $\var{cnt}=582$, the total number of rounds is 10. This process results in the cipher image $\bm{I'}$.
        \end{itemize}

\item \textit{The decryption procedure}:
        \begin{itemize}[leftmargin=*]
             \item \textit{Step 1}:
            Initialize a new counter $\var{cnt}_{\rm d}=0$. This counter increments after each image decryption and resets to zero upon reaching its maximum value.
            
             \item \textit{Step 2}:
            Run network~\eqref{eq:HNNstimuli} for $g(\var{cnt}, T)+1$ iterations and use the final output to reconstruct $N$ from $R'$.
            
              \item \textit{Step 3}:
            Iterate network~\eqref{eq:HNNstimuli} for $2M^2-M \times N -3$ iterations to generate sequences $\bm{F}'$ and $\bm{G}$.
            
             \item \textit{Step 4}:
            Revert $\bm{I}^{**}$ from $\bm{I}'$ using the inverse Cat map. The number of iterations is $4+\sum\nolimits_{i=0}^3\var{cnt_d}[i]2^i$. Subsequently, conduct an XOR operation between the last two elements of $\bm{I}^{**}$ and $\bm{G}$.
            
            \item \textit{Step 5}:
            Retrieve $\var{cnt}$ and compare it with $\var{cnt}_{\rm d}$. If $\var{cnt}$ matches $\var{cnt}_{\rm d}$, one can obtain $\bm{I}^*$ via 
            \begin{equation}
            I^*(i)=G(i) \oplus I^{**}(i).
            \end{equation}
        
            \item \textit{Step 6}:
            Trim the excess pixels in $\bm{I}^*$ to recover the plain image $\bm{I}$.
        \end{itemize}
\end{itemize}

\begin{table}[!htb]
\renewcommand{\arraystretch}{1.3}
 \caption{The method to obtain sequence $\bm{G}$}
 \centering
 %\centering
 \resizebox{\columnwidth}{!}{
  \begin{tabular}{c c c }
   \hline\\[-3mm]
   \multicolumn{1}{l}{Value of $P_1(i)$} & \multicolumn{1}{c}{Value of $P_3(i)$} & \multicolumn{1}{c}{\pbox{20cm}{$G(i)$}} \\[1.6ex] \hline
   0    &  0 & $X(i)$[7:0] $\oplus$ $Z(i)$[15:8]  \\
   0    &  6 & $Y(i)$[9:2] $\oplus$ $Z(i)$[11:4]  \\
   2.5  &  0 & $X(i)$[9:2] $\oplus$ $Y(i)$[11:4]  \\
   2.5  &  6 & $X(i)$[15:8] $\oplus$ $Z(i)$[7:0]  \\ [1.4ex]
\hline
\end{tabular}
}
\label{tab:RNG}
\end{table}

\subsection{Key Space and Computational Complexity}

This encryption scheme effectively utilizes the chaotic properties of AHNN, providing a secure method for image encryption. The process is carefully structured to ensure that the encryption and decryption are consistent and reversible, making it suitable for practical cryptographic applications.

\begin{figure}[!htb]
\centering
\includegraphics[width=\BigOneImW]{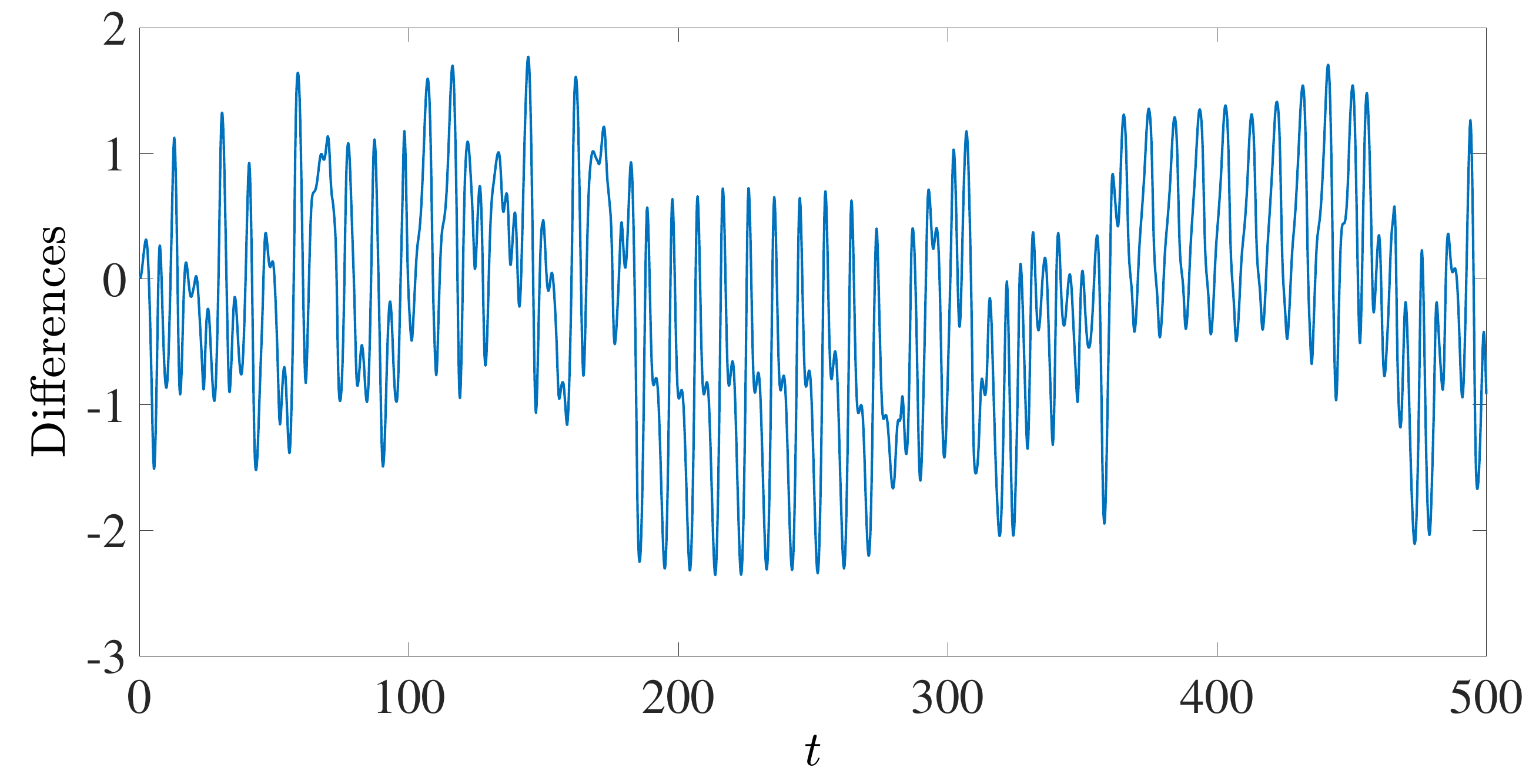}\\
\caption{The differences between two chaotic sequences.} 
\label{fig:diff_x_Ax1}
\end{figure}

The encryption scheme is implemented using Matlab R2018a on the Microsoft Windows 11 operating system. The computational hardware includes an AMD Ryzen 5 5600U CPU and 16 GB of RAM. For ease of testing, the ``quantizer'' function is employed to handle fixed-point number operations. The key space, representing the set of all possible keys that the scheme can utilize for image encryption, is a critical attribute as it determines the scheme's resilience against brute force attacks. In this scheme, the secret key is denoted as $\var{s} \in\{A_1, A_2, A_3, \omega, \omega_1, \omega_2, \omega_3, x_1(0), x_2(0), x_3(0), a, b, c, d, e, T\}$, where $A_i \in [0,10]$ and  $\omega$, $\omega_i$, $x_i(0) \in (0,1)$. The parameters $A_1$, $A_2$, $A_3$, $x_1(0)$, $x_2(0)$, $x_3(0)$ are key parameters with a precision of $10^{-15}$, whereas $\omega, \omega_1, \omega_2, \omega_3$ have a precision of $10^{-7}$. Consequently, the key space size of the scheme is approximately $10^{3\times16} \times 10^{4\times7} \times 10^{3\times15} \times 10^6=10^{127}$. Furthermore, the proposed scheme incorporates a key updating feature, allowing for the encryption of continuous images with varying keys. The counter $\var{cnt}$ plays a pivotal role in this process, as it governs the key updating mechanism. The chaotic sequences generated by the scheme with different counter values are depicted in Fig.~\ref{fig:diff_x_Ax1}.

The encryption computational complexity is another vital aspect. The complexity of encrypting an image of size $M \times N$ includes several processes. Firstly, filling the image to $M \times M$ incurs a complexity of $O(3M^2-3M \times N)$. Secondly, generating chaotic sequences with a length of $M^2$ has a complexity of $O(3M^2)$. Finally, performing confusion and diffusion operations results in a complexity of $O(M^2)$ and $O(S \log_2(M^2))$, respectively, where $S$ is the number of iterations of the Cat map. Therefore, the total complexity is $O(7M^2-3M \times N+S \log_2(M^2))$. For a square image with the size of $M \times M$, the complexity reduces to $O(4M^2+S \log_2(M^2))$.

\begin{figure}[!htb]
 \centering
 \begin{minipage}{\ThreeImW}
  \centering
  \includegraphics[width=\ThreeImW]{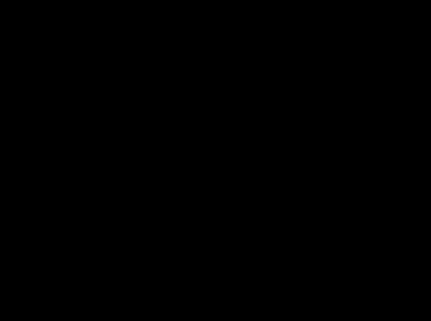} \\
  (a)
 \end{minipage}
 \begin{minipage}{\ThreeImW}
  \centering
  \includegraphics[width=\ThreeImW]{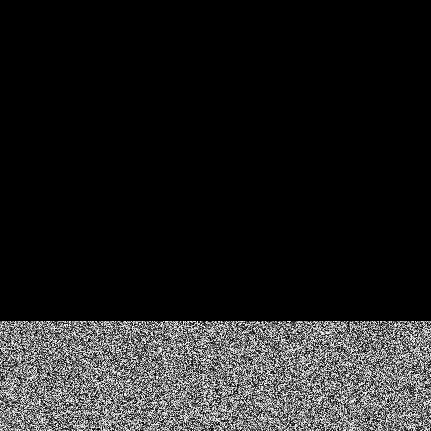}\\
  (b)
 \end{minipage}
 \begin{minipage}{\ThreeImW}
  \centering
  \includegraphics[width=\ThreeImW]{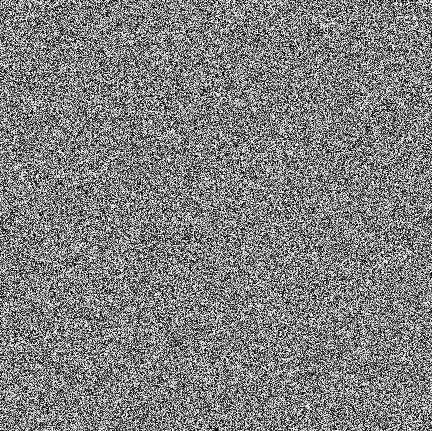}\\
  (c)
 \end{minipage}
 \caption{Encryption flow chart: (a) plain image; (b) filled image; (c) cipher image.}
\label{fig:Fprocess}
\end{figure}

\begin{figure}[!htb]
\centering
    \includegraphics[width=0.8\OneImW]{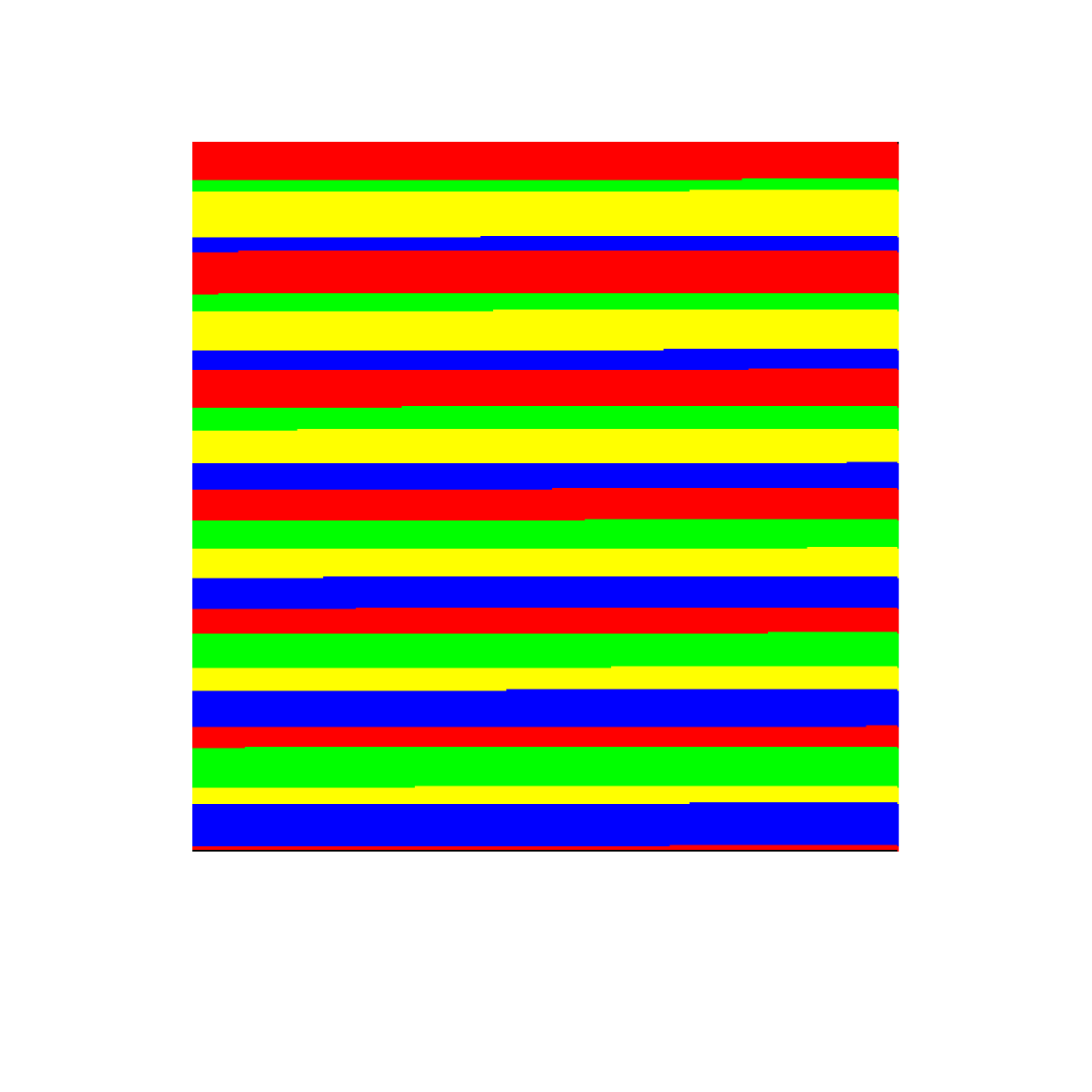}
    \caption{Distribution of four sub-sequences.}
\label{fig:schematic}
\end{figure}

\subsection{Statistical analysis} 

To illustrate the encryption process, a $432 \times 321$ grayscale image with a fixed value of zero is used, as shown in Fig.~\ref{fig:Fprocess}. In this scheme, the chaotic sequence $\bm{G}$, which plays a crucial role in obfuscating the plain images, consists of four sub-sequences. Figure~\ref{fig:schematic} displays the distribution of these sub-sequences, with each corresponding to one of the methods listed in Table~\ref{tab:RNG}. The alternation of methods across the strips enhances the scheme's security. The randomness of sequence $\bm{G}$ is a key indicator of the scheme's security. To test this, the NIST statistical test suite, encompassing 15 different tests, is employed. When the secret key is $[A_1=5, A_2=5, A_3=12, \omega=0.01, \omega_1=0.2, \omega_2=0.22, \omega_3=0.21, x_{1}(0)=0, x_{2}(0)=0.1, x_{3}(0)=0, a=1, b=1, c=1, d=1, e=1, T=5]$, the test results for sequence $\bm{G}$ are summarized in Table~\ref{tab:NIST}, indicating that all bit streams pass the statistical test, as evidenced by P-values higher than the reference value of 0.01.

\begin{table}[!htb]
 \renewcommand{\arraystretch}{1.3}
 \caption{NIST testing results of chaotic sequence $\bm{G}$.}
 \centering
 \resizebox{\columnwidth}{!}{
  \begin{tabular}{l c c }
   \hline\\[-3mm]
   \multicolumn{1}{l}{Statistical tests} & \multicolumn{1}{c}{$k=1.15$} & \multicolumn{1}{c}{\pbox{20cm}{$k=1$}} \\[1.6ex] \hline
   Frequency       & 0.2896   & 0.3190  \\
   Block Frequency	     & 0.7791   & 0.2368  \\
   Cumulative Sums 1$\setminus$2  & 0.4559$\setminus$0.3838 & 0.0102$\setminus$0.3838  \\
   Runs        & 0.5749   & 0.9716  \\
   Longest Run      & 0.5341   & 0.0757  \\
   Rank        & 0.3838   & 0.4190  \\
   FFT        &  0.2354   & 0.9558  \\
   Non-overlapping Template (Average) & 0.6993   & 0.6579  \\
   Overlapping Template    & 0.8831   & 0.4190  \\
   Universal       & 0.1372   & 0.4943  \\
   Approximate Entropy    & 0.1223   & 0.4559  \\
   Random Excursions (Average)  & 0.6371   & 0.3924  \\
   Random Excursions Variant (Average)& 0.5341   & 0.2429  \\
   Serial 1$\setminus$2    & 0.9114$\setminus$0.4190 & 0.6993$\setminus$0.9114  \\
   Linear Complexity     & 0.7791   & 0.1537  \\ [1.4ex]
   \hline
 \end{tabular}
}
\label{tab:NIST}
\end{table}

\begin{figure}[!htb]
\centering
 \begin{minipage}{0.9\twofigwidth}
  \centering
  \includegraphics[width=0.9\twofigwidth]{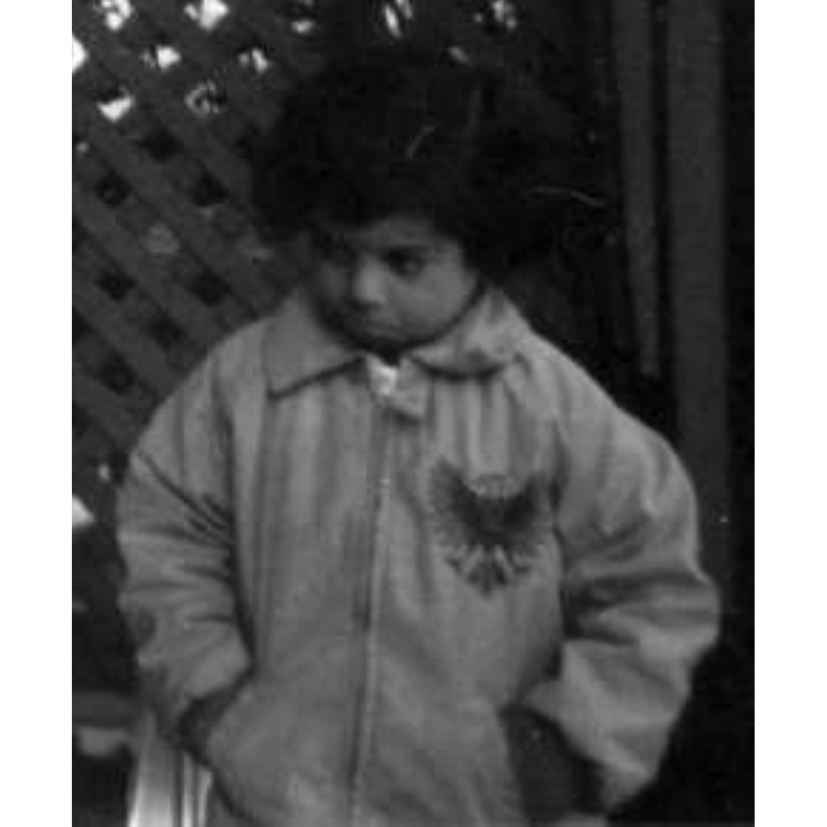}\\
  (a)
 \end{minipage}
 \begin{minipage}{0.9\twofigwidth}
  \centering
  \includegraphics[width=0.9\twofigwidth]{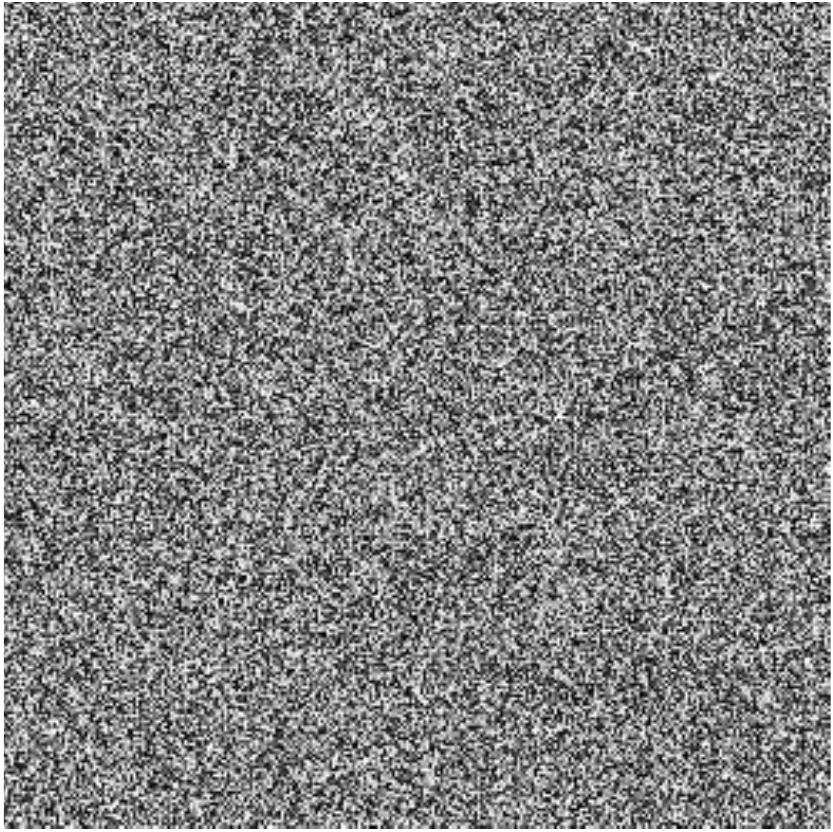}\\
  (b)
 \end{minipage}
 \begin{minipage}{0.9\twofigwidth}
  \centering
  \includegraphics[width=0.9\twofigwidth]{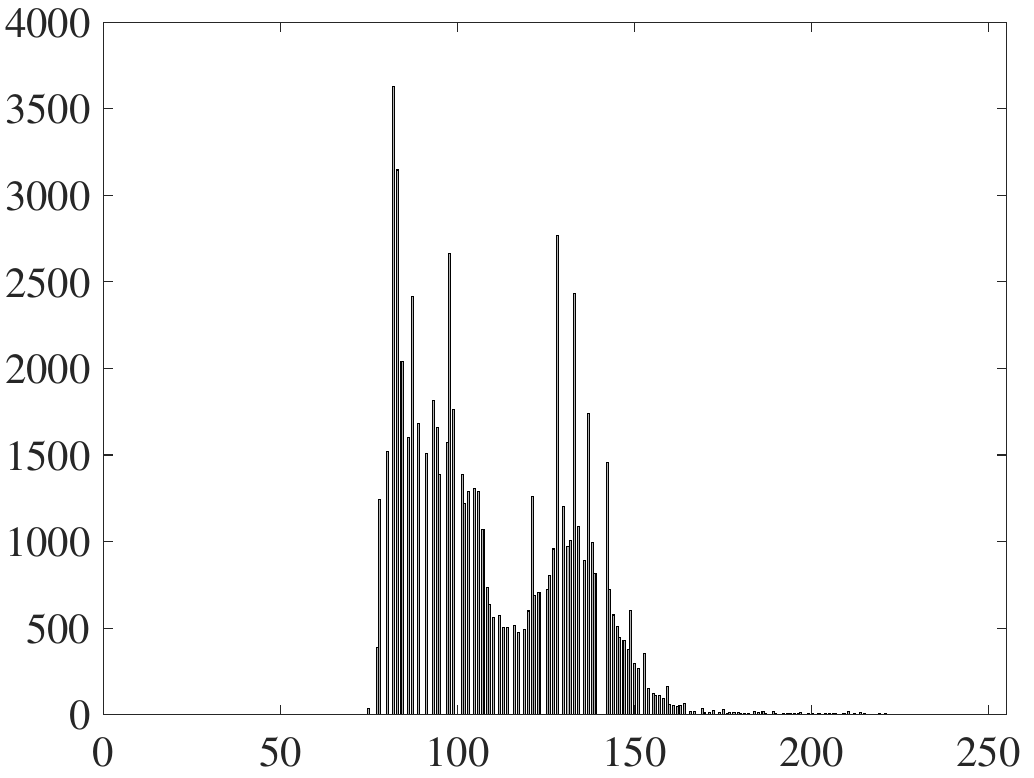}\\
  (c)
 \end{minipage}
 \begin{minipage}{0.9\twofigwidth}
  \centering
  \includegraphics[width=0.9\twofigwidth]{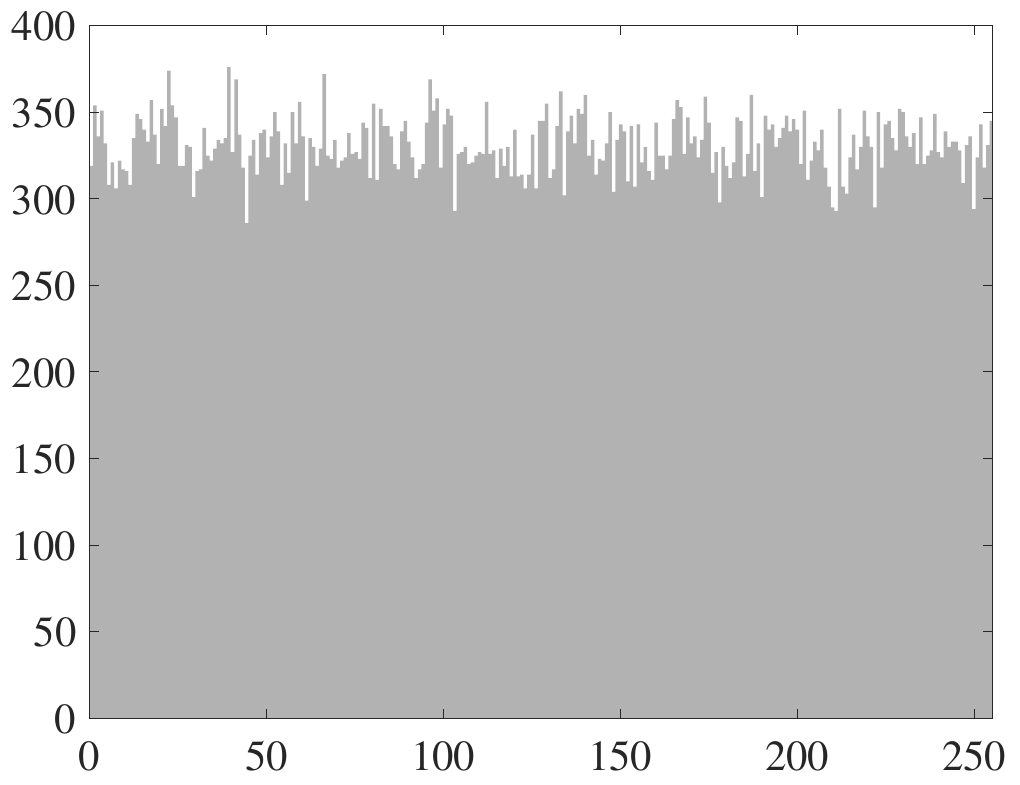}\\
  (d)
 \end{minipage}
 \caption{Histogram analysis: (a) plain image; (b) cipher image; (c) histogram of plain image; (d) histogram of cipher image.}
 \label{fig:Histogram}
\end{figure}

A histogram is a common tool to visually represent the distribution of pixel gray levels in an image. Ideally, the pixel count for different gray levels in the original image should be unevenly distributed. An effective encryption algorithm should ensure that the histogram of the cipher image remains evenly distributed to resist statistical attacks. This can be observed in the histograms of the plain image ``pout.tif", and its encrypted counterpart, as depicted in Fig.~\ref{fig:Histogram}. These histograms show that the pixel values in the plain image are unevenly distributed, and the distribution in the cipher image is relatively uniform, thus indicating the algorithm's resilience against statistical attacks.

\begin{figure}[!htb]
 \centering
 \begin{minipage}{\ThreeImW}
  \centering
  \includegraphics[width=\ThreeImW]{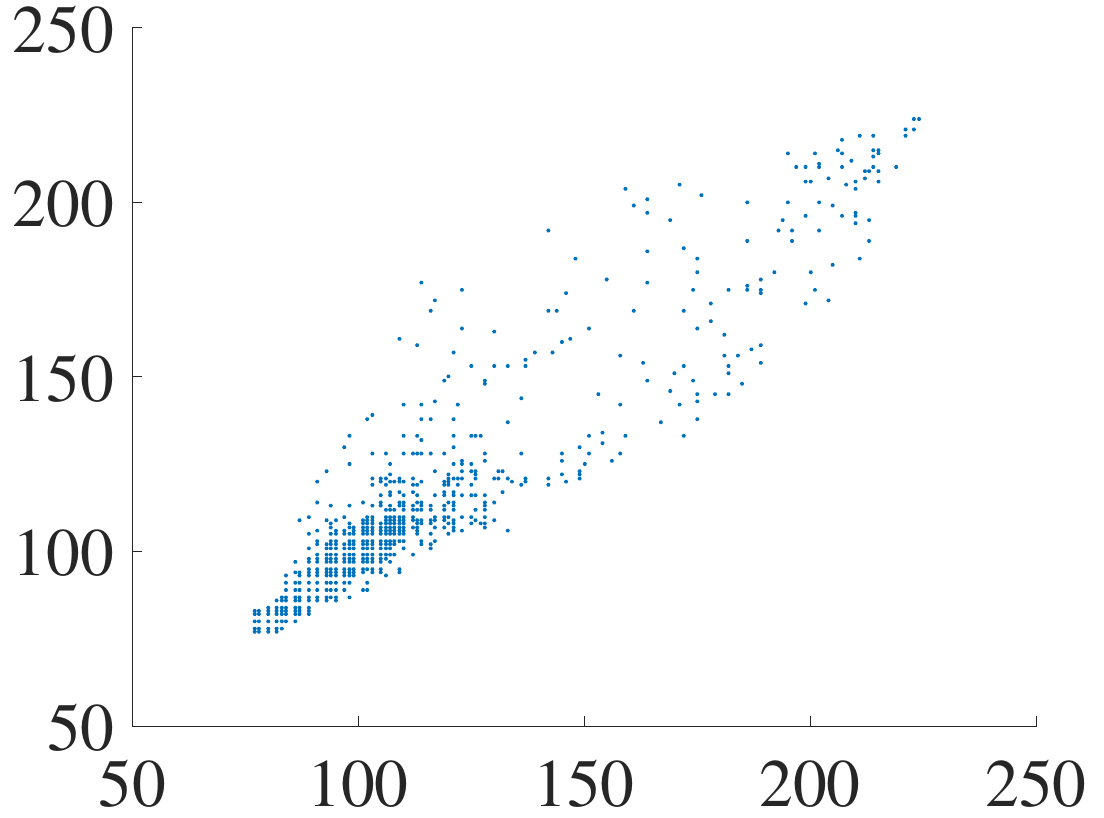}\\
  (a)
 \end{minipage}
 \begin{minipage}{\ThreeImW}
  \centering
  \includegraphics[width=\ThreeImW]{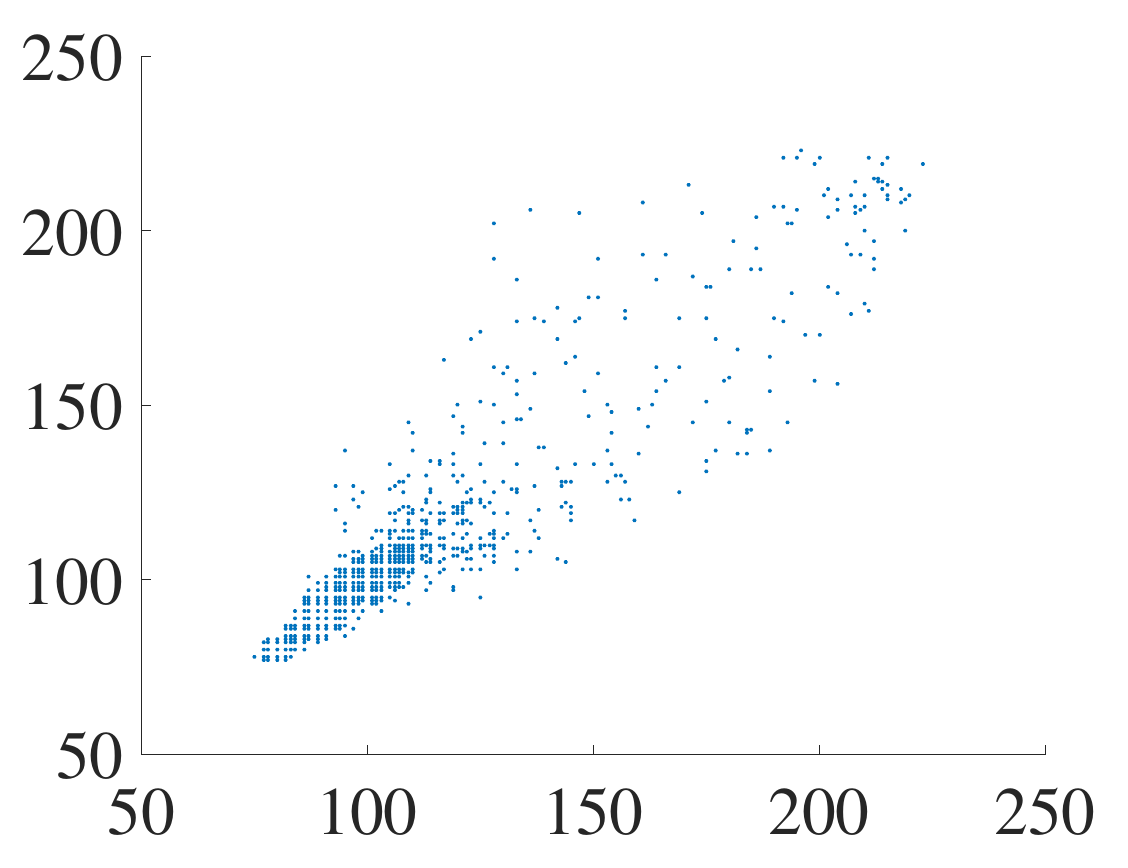}\\
  (b)
 \end{minipage}
 \begin{minipage}{\ThreeImW}
  \centering
  \includegraphics[width=\ThreeImW]{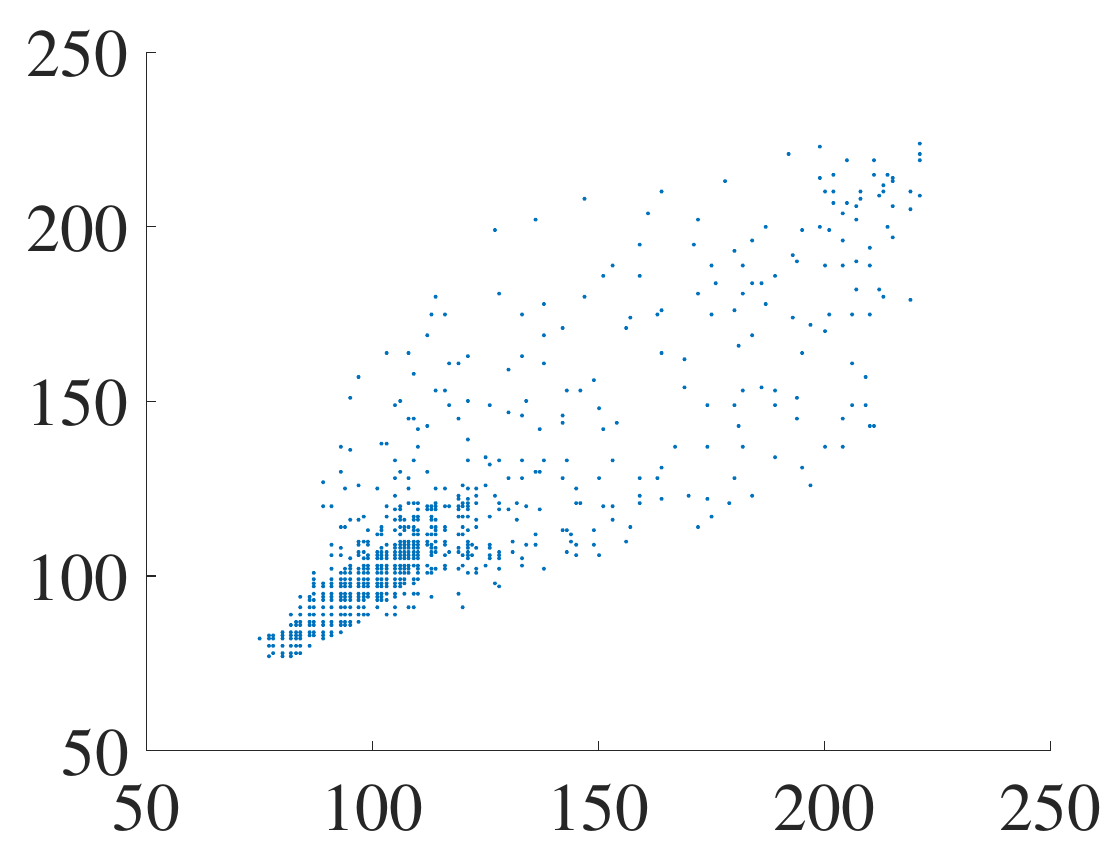}\\
  (c)
 \end{minipage}
 \begin{minipage}{\ThreeImW}
  \centering
  \includegraphics[width=\ThreeImW]{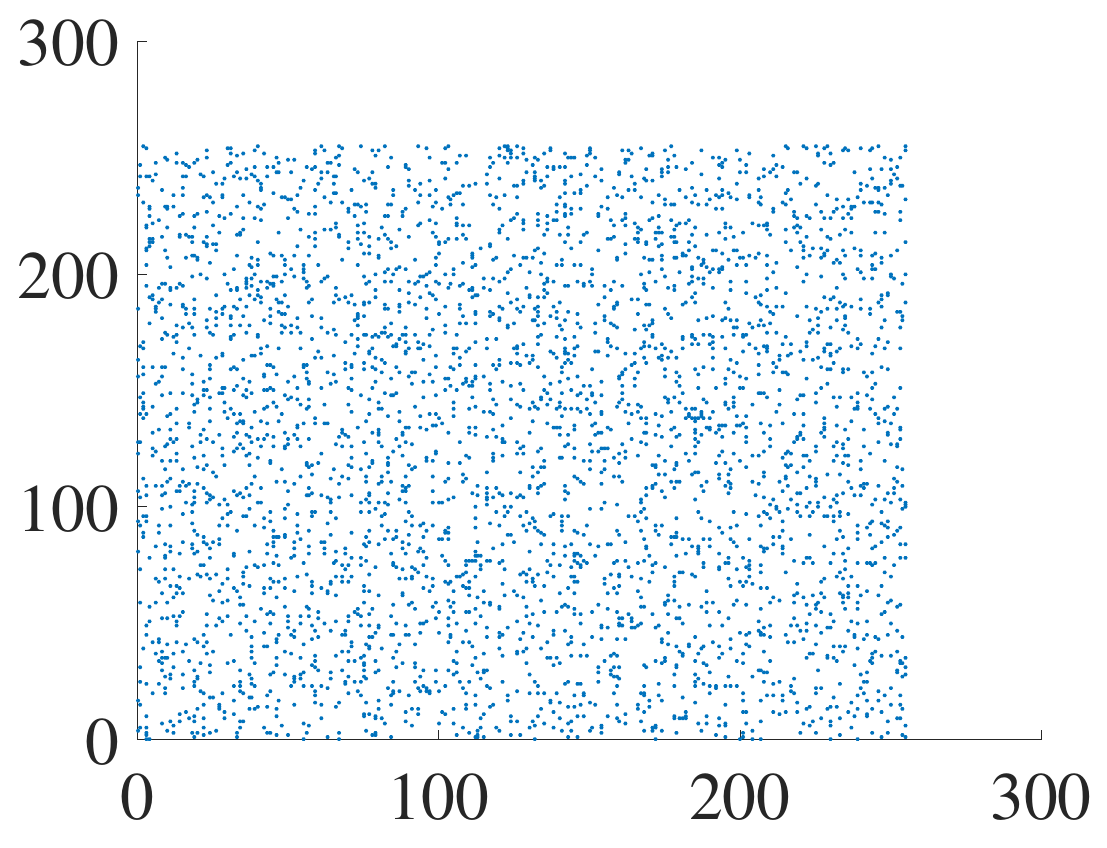}\\
  (d)
 \end{minipage}
 \begin{minipage}{\ThreeImW}
  \centering
  \includegraphics[width=\ThreeImW]{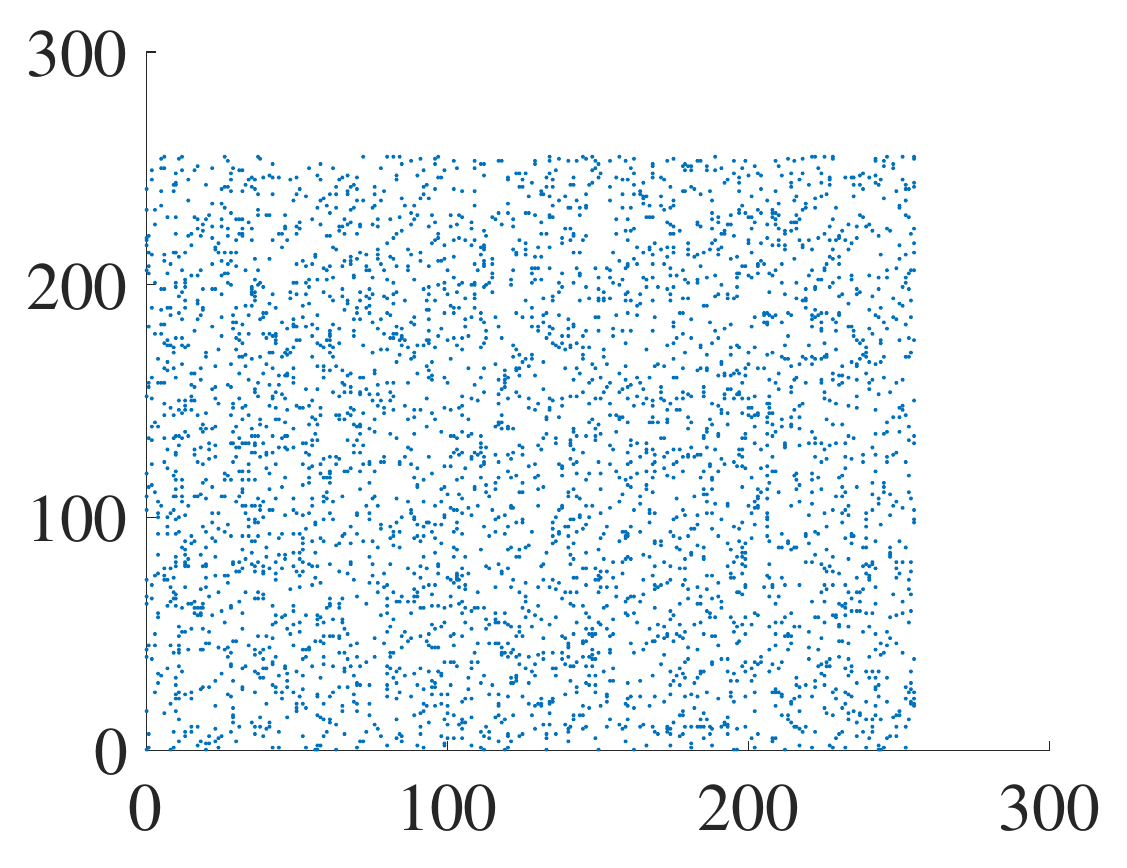}\\
  (e)
 \end{minipage}
 \begin{minipage}{\ThreeImW}
  \centering
  \includegraphics[width=\ThreeImW]{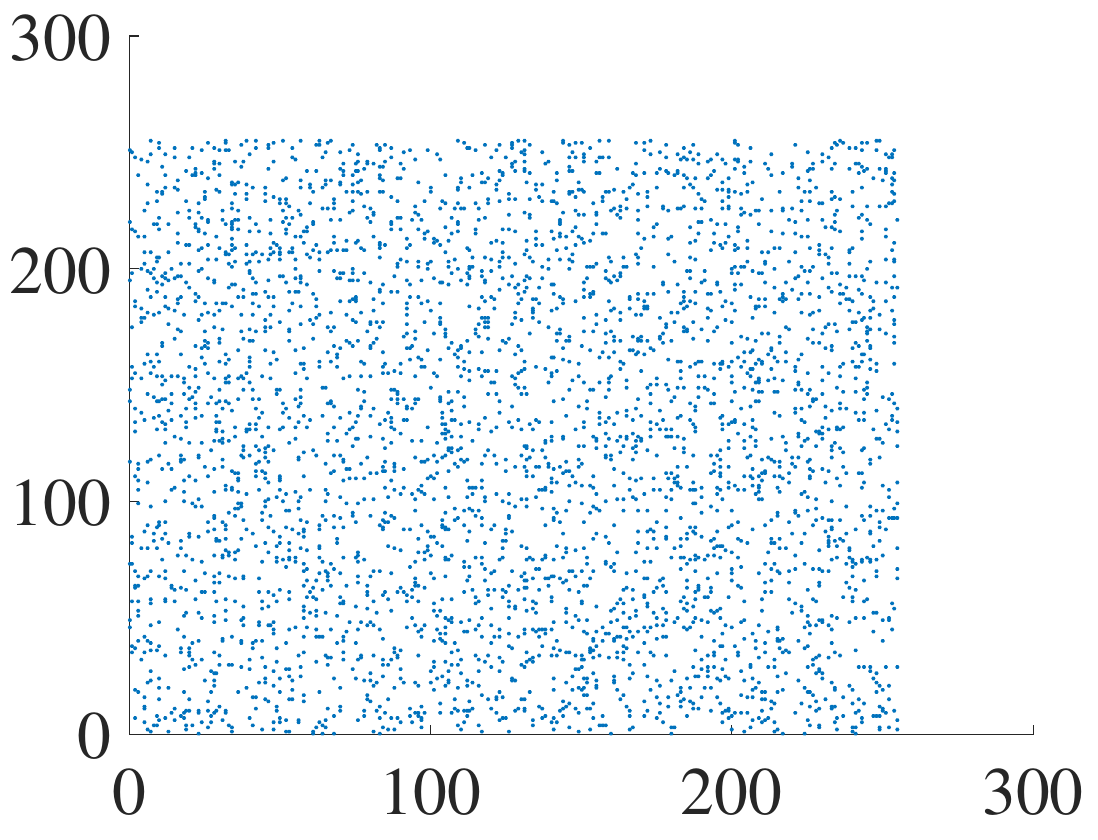}\\
  (f)
 \end{minipage}
\caption{Correlation analysis: (a) horizontal correlation of plain image; (b) vertical correlation of plain image; (c) diagonal correlation of plain image; (d) Horizontal correlation of cipher image; (e) vertical correlation of cipher image; (f) diagonal correlation of cipher image.}
\label{fig:Correlation}
\end{figure}

Correlation distributions of the plain and encrypted images provide additional insight into the effectiveness of the encryption scheme. These distributions, as shown in Fig.~\ref{fig:Correlation}, are obtained by randomly selecting 10,000 pairs of adjacent pixels in horizontal, vertical, and diagonal directions. The correlation coefficients for different types of images are summarized in Table~\ref{tab:Correlation}. The results demonstrate a significantly reduced correlation between adjacent pixels in the encrypted image, enhancing security.

\begin{table}[!htb]
\renewcommand{\arraystretch}{1}
	\caption{Correlation coefficient of different cipher images.}
	\centering
	%\centering
	\resizebox{\columnwidth}{!}{
		\begin{tabular}{c c c c c}
			\hline\\[-3mm]
			\multicolumn{1}{c}{Image} & \multicolumn{1}{c}{Size} & \multicolumn{1}{c}{Horizontal} & \multicolumn{1}{c}{Vertical} & \multicolumn{1}{c}{\pbox{20cm}{Diagonal}} \\[1.6ex] \hline
			cell.tif & 191\texttimes 159 & 0.0017  & 0.0012  & 0.0014 \\
			pout.tif & 240\texttimes 291 & 0.0063  & -0.0060  & -0.0096 \\
			tire.tif & 232\texttimes 205 & 0.0034  & -0.0032  & 0.0043 \\
			lena.tif & 256\texttimes 256 & -0.0011  & 0.0017  & 0.0023 \\
			all-zero & 432\texttimes321 & -0.0036  & 0.0021  & 0.0038 \\
			moon.tif & 358\texttimes 537 & 0.0009  & -0.0018  & -0.0013 \\	
			all-255  & 1280\texttimes 720 & -0.0028  & 0.0019  & -0.0027 \\
			 [1.4ex]
			\hline
		\end{tabular}
	}
\label{tab:Correlation}
\end{table}

Furthermore, three additional metrics are introduced to assess the security of the encryption scheme. Information entropy serves as a unique method to sum up the probability of every output from the signal source, providing a quantitative measure of the randomness of image information. It is
\begin{equation}
H(I)=-\sum_{n=0}^{255}Pr(I_i)\log_2Pr(I_i),
\end{equation}
where $\var{Pr}(I_i)$ represents the occurrence probability of the $i$-th pixel. It is widely accepted that the randomness of the signal source is better if the information entropy is closer to eight.

To evaluate the disparity between the plain image and the cipher image, the Mean Square Error $(\var{MSE})$ is
\begin{equation}
\var{MSE}=\frac{1}{N^2}\sum_{i=0}^{255}\sum_{j=0}^{255}(I(i, j)-I'(i, j))^2,
\end{equation}
where $I(i, j)$ and $I'(i, j)$ denote the pixels in the $i$-th row and $j$-th column of the plain and cipher images' gray matrices, respectively, and $N^2$ is the total number of pixels in the images. A high $\var{MSE}$ value indicates a substantial deviation of the encryption result from the original image.

Additionally, the Peak Signal-to-Noise Ratio $(\var{PSNR})$ is  
\begin{equation}
\var{PSNR}=20\log_{10}\left(\frac{255}{\sqrt{\var{MSE}}}\right),
\end{equation}
to measure the impact of image compression. This index is inversely related to $\var{MSE}$. The test results of these three indices for seven different types of images are presented in Table~\ref{tab:Entropy}, demonstrating the high security level of the proposed encryption scheme.

\begin{table}[!htb]
    \renewcommand{\arraystretch}{1.3}
    \caption{Comparison of cipher images in terms of three objective metrics.}
    \centering
    \resizebox{\columnwidth}{!}{
        \begin{tabular}{c c c c c}
            \hline
            \multicolumn{1}{c}{Image} & \multicolumn{1}{c}{Size} & \multicolumn{1}{c}{Entropy} & \multicolumn{1}{c}{$\text{MSE}$} & \multicolumn{1}{c}{$\text{PSNR}$} \\
            \hline
            cell.tif & 191$\times$159 & 7.9951  & 6636.4994  & 9.9454 \\
            tire.tif & 232$\times$205 & 7.9968  & 14409.1781 & 6.5784 \\
            pout.tif & 240$\times$291 & 7.9979  & 7095.0603  & 9.6552 \\
            lena.tif & 256$\times$256 & 7.9977  & 7754.3216  & 9.2693 \\
            all-zero & 432$\times$321 & 7.9990  & 18929.5313 & 5.3934 \\
            moon.tif & 358$\times$537 & 7.9993  & 15021.7248 & 6.3976 \\	
            all-255 & 1280$\times$720 & 7.9999  & 16956.1530 & 5.8715 \\
            \hline
        \end{tabular}
    }
    \label{tab:Entropy}
\end{table}

\subsection{Ability against tradition attack}

Differential attacks, which analyze the differences between cipher images derived from slightly varied plaintexts, are effective strategies for compromising cryptosystems. To quantify these differences, metrics such as the Number of Pixel-Changing Rates ($\text{NPCR}$) and the Unified Average Changed Intensity ($\text{UACI}$) are employed. These are 
\begin{equation}
\text{NPCR}=\frac{1}{MN}\sum_{h=1}^{M}\sum_{j=1}^{N}W(h, j)\times100\%
\end{equation}
and
\begin{equation}
\text{UACI}=\frac{1}{MN}\sum_{h=1}^{M}\sum_{j=1}^{N}\frac{|I_1(h, j)-I_2(h, j)|}{255}\times100\%,
\end{equation}
where $I_1(h, j)$ and $I_2(h, j)$ represent the pixel values of two cipher images
obtained from the corresponding plain images with only a one-bit difference, and 
\begin{equation}
W(h, j)=
   \begin{cases}
       0 & \mbox{if }  I_1(h, j)=I_2(h, j); \\
       1 & \mbox{otherwise}.
   \end{cases}
\end{equation}
The results for these metrics are 99.6173\% and 33.5303\%, respectively, closely aligning with the ideal values of 99.6094\% and 33.4635\%.

\begin{figure}[!htb]
 \centering
 \begin{minipage}{\twofigwidth}
  \centering
  \includegraphics[width=0.9\twofigwidth]{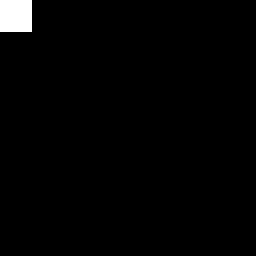}\\
  (a)
 \end{minipage}
 \begin{minipage}{\twofigwidth}
  \centering
  \includegraphics[width=0.9\twofigwidth]{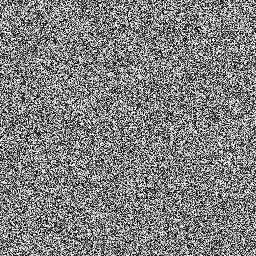}\\
  (b)
 \end{minipage}
 \caption{Result of chosen-plaintext attack:
 (a) $I_1 \oplus I_2$; (b) $C_1 \oplus C_2$.}
\label{fig:plaintext}
\end{figure}

In a chosen-plaintext attack scenario, the attacker selects a plain image and obtains the corresponding cipher one. The experiment involves two chosen images: $I_1$, a constant image, and $I_2$, a slightly altered version of $I_1$. Their resulting cipher images are $C_1$ and $C_2$, respectively. The encryption scheme is vulnerable if $I_1 \oplus I_2=C_1 \oplus C_2$. However, as Fig.~\ref{fig:plaintext} demonstrates, the proposed scheme effectively counters this attack through its key updating mechanism, which encrypts consecutive images using different keys. This ability and other metrics are compared with some counterparts as shown in Table~\ref{tab:compare}.

\begin{table}[!htb]
\renewcommand{\arraystretch}{1.3}
\caption{Comparison between the proposed scheme with counterparts in terms of typical metrics.} 
 \centering
 \resizebox{\columnwidth}{!}{
 \begin{threeparttable}
  \begin{tabular}{c c c c c} 
   \hline\\[-3mm]
   \multicolumn{1}{l}{Schemes} & \multicolumn{1}{c}{Information entropy} & \multicolumn{1}{c}{$\text{NPCR}$} & \multicolumn{1}{c}{$\text{UACI}$}  & \multicolumn{1}{c}{Key updating} \\[1.6ex] \hline
   \cite{Yu:Hopfield:TNSE23}         &  7.9977 & 99.6078 & 33.4875 & $\times$\\
   \cite{Li:encryption:TNNLS2021}    &  7.9973 & 99.5700 & 33.5000 & $\times$\\
   \cite{njitacke:Tabu-analog:TII22} &  7.9980 & 99.7300 & 33.5765 & $\times$\\
   \cite{zhang:image:chaos21}        &  7.9974 & NR\tnote{1} & NR & $\times$\\
   \cite{Lin:FPGAchaos:TIE2021}      &  7.9976 & NR & NR & $\times$\\
   This work                         &  7.9977 & 99.6173 & 33.5303 & $\checkmark$ \\ [1ex]
   \hline
  \end{tabular}
  \begin{tablenotes}
        \footnotesize
        \item[1] Not reported.
   \end{tablenotes}
\end{threeparttable}
}
\label{tab:compare}
\end{table}

\begin{figure}[!htb]
 \centering
 \begin{minipage}{\twofigwidth}
  \centering
  \includegraphics[width=0.9\twofigwidth]{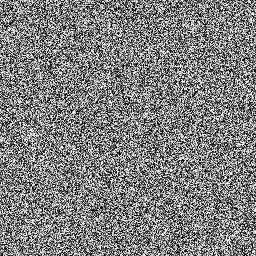}\\
  (a)
 \end{minipage}
 \begin{minipage}{\twofigwidth}
  \centering
  \includegraphics[width=0.9\twofigwidth]{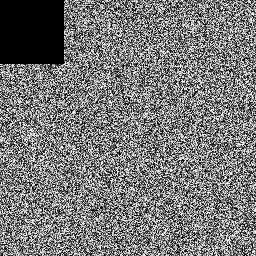}\\
  (b)
 \end{minipage}
  \begin{minipage}{\twofigwidth}
  \centering
  \includegraphics[width=0.9\twofigwidth]{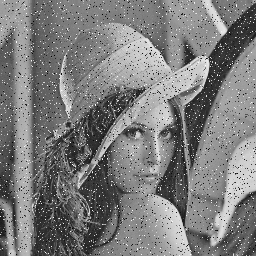}\\
  (c)
 \end{minipage}
 \begin{minipage}{\twofigwidth}
  \centering
  \includegraphics[width=0.9\twofigwidth]{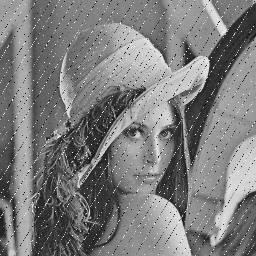}\\
  (d)
 \end{minipage}
 \caption{Robustness analysis:
 (a) cipher image with noise; (b) cipher image with data loss; (c) deciphered image of (a); (d) deciphered image of (b).}
\label{fig:noise}
\end{figure}

Noise attacks involve data loss or modification of the cipher image during transmission or storage. A robust encryption scheme should be able to accurately decrypt even corrupted images. To evaluate this, the proposed scheme was tested with a cipher image subjected to 10\% salt \& pepper noise and 6\% data loss, as shown in Fig.~\ref{fig:noise}(a) and (b). The deciphered images, presented in Fig.~\ref{fig:noise}(c) and (d), indicate the scheme's robustness against noise attacks.

\section{Conclusion}

This study explored the enhancement of the dynamics of the HNN through the application of time-variant external stimuli.
Four distinct AHNN configurations, each with a unique topology, demonstrated that appropriate stimuli application does not compromise HNN stability but significantly broadens the scroll number of attractors. This method effectively controlled the nonlinear system's dynamics, enhancing its complexity and applicability in various scientific and engineering fields. Subsequently, an image encryption algorithm based on AHNN was developed. This algorithm, incorporating dynamic key changes with time-variant stimuli, ensured robust security in multimedia communications. The time-variant stimulus approach presented herein provided valuable insights for adjusting dynamics in nonlinear systems.

% References
% \bibliographystyle{Bibliography/IEEEtranTIE}
\bibliographystyle{IEEEtran_doi}
%\IEEEtriggeratref{49}
\bibliography{Bibliography/IEEEabrv, Bibliography/References}\ %IEEEabrv instead of full

\vspace{-1.2cm}
\begin{IEEEbiography}[{\includegraphics[width=1in,height=1.25in,clip,keepaspectratio]{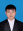}}]
{Xuenan Peng} received his B.S. degree from Shaoyang College in 2018 and is currently pursuing his Ph.D. at the School of Mathematics and Computational Science, Xiangtan University, Xiangtan, China. His current research interests include multi-scroll
systems and circuits, memristive systems and circuits, chaos and fractional-order chaotic systems and circuits, chaos-based image encryption schemes, and conservative chaotic systems.
\end{IEEEbiography}

\vskip 0pt plus -1fil

%\vspace{-1cm}

\begin{IEEEbiography}[{\includegraphics[width=1in,height=1.25in,clip,keepaspectratio]{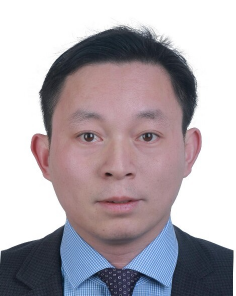}}]
{Chengqing Li} (M'07-SM'13) obtained his M.Sc. degree in Applied Mathematics from Zhejiang University, China, in 2005, followed by a Ph.D. in Electronic Engineering from the City University of Hong Kong in 2008. He served as a Post-Doctoral Fellow at The Hong Kong Polytechnic University until September 2010. Subsequently, he joined the College of Information Engineering at Xiangtan University, China. From April 2013 to July 2014, he worked at the University of Konstanz, Germany, supported by the Alexander von Humboldt Foundation. Since April 2018, he has been a Professor at the School of Computer Science and Electronic Engineering, Hunan University, China, and currently holds the position of Full Professor at the School of Computer Science, Xiangtan University, China. His research primarily focuses on the dynamics analysis of digital chaotic systems and their applications in multimedia security. Over the past 20 years, he has published over 60 papers in this area, receiving more than 5600 citations and achieving an h-index of 40. He is a Fellow of the IET.
\end{IEEEbiography}

\vskip 0pt plus -1fil

\begin{IEEEbiography}[{\includegraphics[width=1in,height=1.25in,clip,keepaspectratio]{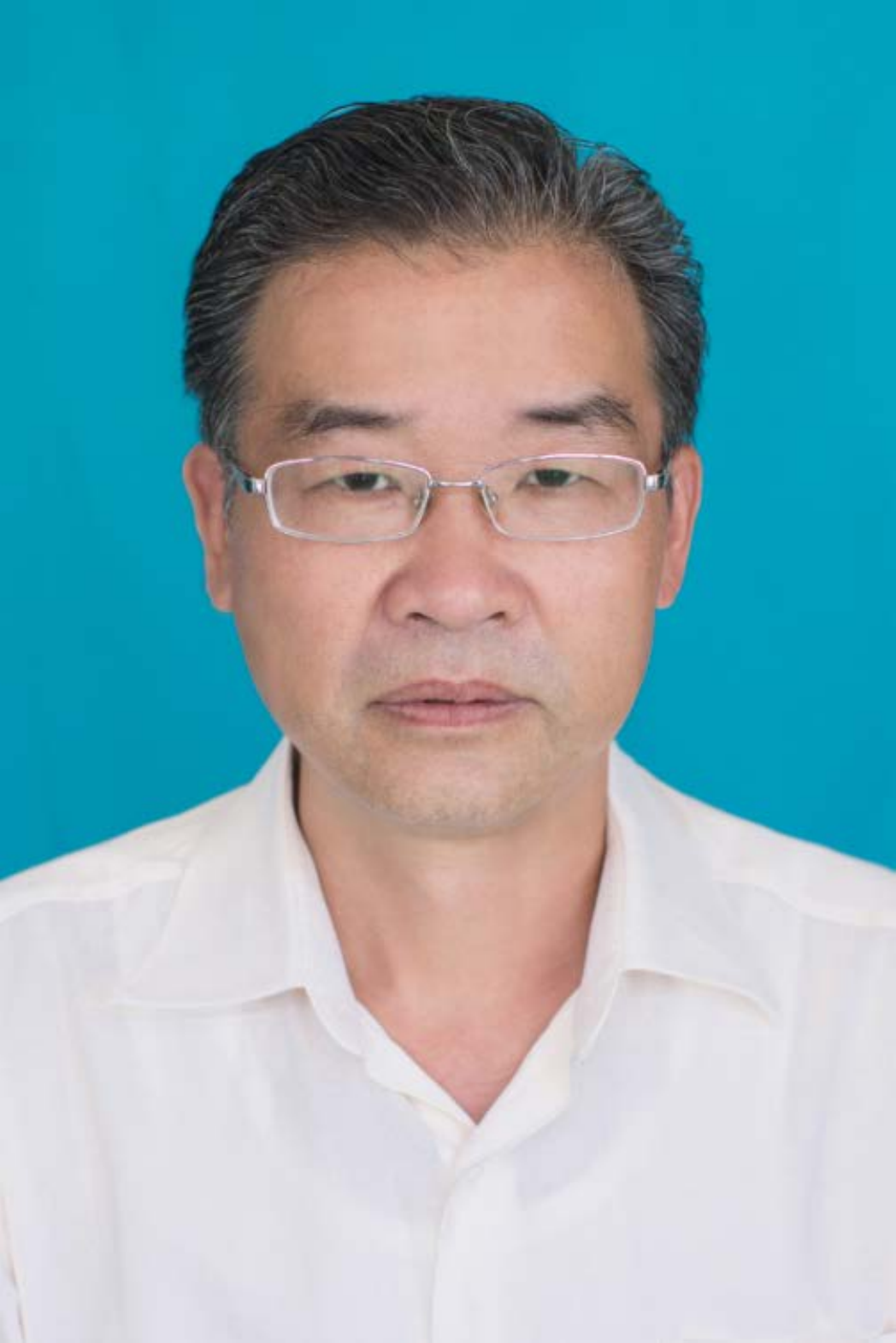}}]
{Yicheng Zeng} completed his undergraduate studies at Huaihua College in 1982, followed by a Master's degree from Sichuan Normal University in 1991 and a Ph.D. from Zhejiang University in 2002. Since 2004, he has been serving as a Professor of Electronics at the School of Physics and Optoelectronic Engineering, Xiangtan University. He was the initiator of this work and passed away in December 2022.
\end{IEEEbiography}

%If you want them closer to one another, adjust 0pt to something like -2\baselineskip.
\vskip -2\baselineskip plus -1fil

\begin{IEEEbiography}[{\includegraphics[width=1in,height=1.25in,clip,keepaspectratio]{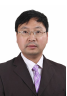}}]
{Chunlai Li} earned his Ph.D. from the College of Automation at Guangdong University of Technology, Guangzhou, China, in 2012. He currently holds a professorship at the School of Computer Science, Xiangtan University, China. His research interests are diverse, encompassing nonlinear circuits, chaos dynamics, memristors, memristive systems, image encryption techniques, neuron model development, and the dynamics of neural networks.
\end{IEEEbiography}

\end{document}